\documentclass[a4paper,fleqn]{cas-dc}

\usepackage[table]{xcolor}
\usepackage{booktabs}
\usepackage{tabularx}
\usepackage{array}
\usepackage[table]{xcolor}
\usepackage{booktabs}
\usepackage{tabularx}
\usepackage{array}

\definecolor{PanelGray}{RGB}{232,235,236}
\definecolor{HeaderGray}{RGB}{218,222,223}
\definecolor{LightGray}{RGB}{247,247,247}

\definecolor{PanelGray}{RGB}{232,235,236}
\definecolor{HeaderGray}{RGB}{218,222,223}
\definecolor{PriorGray}{RGB}{247,247,247}
\definecolor{OursGray}{RGB}{230,233,234}

\usepackage[table]{xcolor}
\usepackage{booktabs}
\usepackage{tabularx}
\usepackage{array}
\usepackage[authoryear,longnamesfirst]{natbib}
\setcitestyle{super,open={},close={},citesep={,}}
\def\tsc#1{\csdef{#1}{\textsc{\lowercase{#1}}\xspace}}
\tsc{WGM}
\tsc{QE}

\begin{document}
\let\WriteBookmarks\relax
\def\floatpagepagefraction{1}
\def\textpagefraction{.001}

\shorttitle{}    

\shortauthors{Zhou et al.}  

\title [mode = title]{PeriaTrace: Tooth-Level Grounding Diagnosis and Patient-Facing Explanations from Periapical Radiographs}  


%

\author[1,2]{Zhenhuan Zhou}[orcid=0009-0000-2187-7184]

\affiliation[1]{organization={College of Computer Science, Nankai University},
            city={Tianjin},
            postcode={300350}, 
            country={China}}   

\affiliation[2]{organization={Key Laboratory of Data and Intelligent System Security, Ministry of Education},
            city={Tianjin},
            postcode={300350}, 
            country={China}}   






\author[3]{Yuchen Zhang}[orcid=0009-0005-5448-3292]






\affiliation[3]{organization={Department of stomatology, Tianjin Union Medical Center, The First Affiliated Hospital of Nankai University, China},
            city={Tianjin},
            postcode={300121}, 
            country={China}}

\cormark[1]
\ead{zhangyuchen@umc.net.cn}

\author[4]{Jingbo Zhu}[orcid=0009-0006-6536-6822]

\affiliation[4]{organization={The College of Software of Nankai University},
            city={Tianjin},
            postcode={300350}, 
            country={China}}

\author[5]{Shiqing Ma}[orcid=0000-0001-9002-4054]

\affiliation[5]{organization={Department of Stomatology, The Second Hospital of Tianjin Medical University},
            city={Tianjin},
            postcode={300211}, 
            country={China}}

\author[1,2]{Yanlin Wu}[orcid=0000-0002-2087-275X]
        
\author[1,2]{Tao Li}[orcid=0000-0002-1273-0487]

\cormark[1]
\ead{litao@nankai.edu.cn}


\credit{}     

\cortext[1]{Corresponding author}



\begin{abstract}
Periapical radiographs (PRs) are essential for dental diagnosis, particularly in endodontic practice, as they provide detailed views of local dental and periodontal structures and pathological changes. However, their interpretation and reporting remain time-consuming and require specialized expertise. The technical terminology of conventional radiographic reports also limits patient understanding of diagnostic findings and recommended care, while existing dental report-generation methods offer limited fine-grained visual grounding, making it difficult to link specific diagnoses to the corresponding teeth and radiographic evidence. Here we present PeriaTrace, a three-stage framework integrating tooth-level diagnosis, evidence-grounded report generation, and patient-facing communication from PRs. PeriaTrace was developed and evaluated using 12,877 PRs from an internal dataset and two additional clinical cohorts. In the internal evaluation, PeriaTrace achieved a macro Dice coefficient of 85.35\%, a BLEU-4 score of 0.6504, an F1 score of 94.37\% for joint tooth localization and Fédération Dentaire Internationale (FDI) identification, and a grounding-tag F1 score of 77.93\%. In external clinical evaluations, both senior endodontists considered 91.16\% and 84.52\% of the generated reports at the two respective cohorts suitable for retention and revision without complete rewriting. Automated report generation required an average of 4.41 seconds per PR, representing a speed-up of several tens of times compared with manual report drafting. In a survey of 196 participants without medical backgrounds, 96.4\% preferred report formats incorporating patient-friendly explanations and care recommendations, while 84.7\% found visual localization cues helpful. These findings demonstrate the potential of PeriaTrace to facilitate efficient, visually traceable grounding dental reporting while supporting clinical review and patient-centred communication.
\end{abstract}


\maketitle

\section{Introduction}\label{Introduction}
Oral diseases are among the most prevalent noncommunicable conditions worldwide, affecting nearly 3.7 billion people and creating a substantial burden for individuals, health systems and societies \cite{who2025oral,peres2019oral}. This burden spans dental caries, periodontal disease, endodontic infection, tooth loss and other conditions that can cause pain, impaired oral function and reduced quality of life \cite{peres2019oral,wen2022global,sarsam2025role}. Beyond symptoms, oral diseases also generate substantial treatment costs and productivity losses, with dental expenditures varying considerably across countries and income settings \cite{jevdjevic2025global}. Timely diagnosis and disease monitoring are therefore central to preventive dentistry, routine clinical care and broader public health strategies \cite{who2022globaloral,peres2019oral}. In this context, imaging-based assessment is not only a technical step in dental diagnosis, but also an important component of clinical decision-making that shapes how disease is detected and treatment is planned \cite{sarsam2025role}, while the communication of imaging findings can influence how patients understand their oral health status \cite{park2024patient}.

In endodontic and restorative dentistry, periapical radiographs (PRs) are the most commonly used imaging modality for assessing dental and periapical abnormalities that cannot be fully evaluated through visual examination alone. Their accessibility, relatively low cost and radiation exposure, together with their ability to provide detailed information on individual teeth, root morphology, periapical bone structures and localized lesions, make them an essential component of routine clinical practice \cite{sarsam2025role}. However, accurate interpretation of these images requires considerable clinical expertise and experience. Clinicians must recognize subtle radiographic signs, accurately associate abnormalities with individual teeth and translate localized imaging findings into clinically meaningful diagnostic reports. This process can be time-consuming and subject to interobserver variability \cite{meusburger2023reliability}. Moreover, conventional radiographic reports often contain specialized terminology that patients may struggle to understand or recall, potentially leading to confusion, anxiety or insufficient awareness of the need for appropriate follow-up care \cite{park2024patient,chen2026,vanDerMee2025formats}. Without further explanation from clinicians, patients may have difficulty understanding their conditions and the corresponding treatment recommendations. To address these challenges, dental artificial intelligence (AI) systems have emerged. Such systems offer opportunities to reduce clinicians' reporting workload while providing patients with more accessible and targeted information about their oral health.

Recent studies have begun to extend dental AI beyond task-specific detection and classification towards more integrated image interpretation and report-oriented systems. Early studies established deep-learning pipelines for automated tooth detection and numbering in panoramic radiographs. Estai et al. combined region-of-interest extraction, object detection and tooth classification to assign Fédération Dentaire Internationale (FDI) numbers to individual teeth, whereas Yüksel et al. developed DENTECT, which simultaneously identifies teeth according to the FDI system and detects five categories of dental treatments \cite{estai2022deep,yuksel2021dental}. DENTEX subsequently established a hierarchical benchmark linking quadrant localization, tooth enumeration and diagnostic labels, enabling abnormal teeth to be identified together with their associated conditions \cite{hamamci2023dentex}. Subsequent work has integrated anatomical localization with pathological assessment. Zhou et al. developed a dual-labeled dataset and fusion model that combines tooth instance segmentation, numbering and condition assessment, with outputs including visualizations and automatically generated reports \cite{zhou2024duallabeled}. Similarly, van Nistelrooij et al. combined the OdontoAI and DENTEX datasets to jointly predict tooth segmentation, FDI numbering and multiple diagnoses, including caries, impacted teeth and periapical lesions \cite{nistelrooij2024combining}. In a complementary approach, Silva et al. developed a semi-supervised framework that combines tooth instance segmentation with large language model-based extraction of disease labels from paired clinical reports. By linking the extracted conditions to segmented tooth instances, their approach enables tooth-level classification across multiple dental conditions without requiring directly annotated tooth-level disease labels \cite{silva2025holistic}. 

Cone-beam computed tomography (CBCT) has become an important modality in dental AI research, with applications spanning adult and pediatric populations and a wide range of tasks, including anatomical segmentation, tooth identification, pathological assessment, and diagnostic reporting \cite{cui2022fully,hu2024fully,zhou2024nkut,bolelli2026multi,zhou2025deep}. Cui et al. developed a fully automated system for individual tooth and alveolar bone segmentation, evaluated on 4,938 CBCT scans from 15 clinical centres, demonstrating its robustness across diverse anatomical conditions and imaging protocols \cite{cui2022fully}. Extending automated analysis to pediatric dentistry, Hu et al. proposed a deep learning framework for three-dimensional segmentation and fine-grained identification of mixed dentition, addressing the anatomical complexity of primary and developing permanent teeth \cite{hu2024fully}. More recently, Zhou et al. integrated tooth segmentation and numbering with disease assessment to generate structured diagnostic reports from CBCT images, further extending automated analysis towards clinical reporting \cite{zhou2025deep}. The ToothFairy2 benchmark has also advanced CBCT research by providing expert annotations of 42 dental and maxillofacial structures for standardized evaluation \cite{bolelli2026multi}.

The development of dental multimodal large language models (MLLMs) has further broadened the scope of automated interpretation. Hao et al. introduced MMOral, a large-scale instruction dataset comprising 20,563 annotated panoramic radiographs and approximately 1.3 million instruction-following instances covering attribute extraction, report generation, visual question answering and image-grounded dialogue. Using this resource, they developed OralGPT through supervised fine-tuning of a vision–language model (VLM) for dental image understanding \cite{hao2025mmoral}. Subsequent work from the same research group has extended this line of research in two directions. OralGPT-Plus incorporates tool use and reinforcement learning to support iterative image reinspection and symmetry-aware diagnostic reasoning, supported by the DentalProbe dataset and the region-annotated MMOral-X benchmark \cite{fan2026oralgptplus}. OralGPT-Omni broadens the scope to multiple dental imaging modalities and diagnostic tasks, including abnormality diagnosis on PRs, and introduces clinically informed reasoning supervision through TRACE-CoT alongside the MMOral-Uni benchmark \cite{hao2026oralgptomni}. In parallel, DentFound employs an instance-guided vision–language architecture for panoramic radiograph interpretation and report generation. Trained and evaluated using multicentre cohorts encompassing more than 101,000 patients, 98 disease categories and 11 post-treatment categories, DentFound demonstrates the potential of instance-level visual guidance to support comprehensive dental reporting and clinical assessment \cite{zhu2026dentfound}. Meng et al. developed DentVLM, a dental-specific VLM trained on large-scale multimodal data to support 36 diagnostic tasks across seven dental imaging modalities, demonstrating its potential for comprehensive dental diagnosis and clinical decision support \cite{meng2026multimodal}.

Despite advances in dental AI, important limitations remain in the comprehensive analysis and clinical application of PRs. A recent systematic review of 47 studies highlighted persistent challenges, including reliance on cropped images, insufficient external validation, and methodological bias \cite{tasdemir2026diagnostic}. First, publicly available PR datasets remain fragmented, primarily supporting individual tasks such as segmentation, detection, or classification. To our knowledge, no publicly available PR dataset simultaneously provides pixel-level segmentation annotations, diagnostic labels, clinician-authored reports, and explicit tooth-level associations between textual findings and visual evidence. Consequently, existing studies often address isolated tasks, while comprehensive reporting workflows supported by PR-specific multimodal annotations remain scarce, limiting the development and generalizability of integrated systems. Second, explicit quality control (QC) for uncertain tooth identification, uncertainty estimation, systematic linkage of diagnostic findings to individual teeth and localized image evidence, and structured reporting with verifiable content remain insufficiently integrated. Beyond dentistry, region-guided and localization-grounded models have linked radiological descriptions to anatomical regions or lesions \cite{tanida2023rgrg,bannur2024maira2}, demonstrating the potential of visual grounding for grounding diagnostic reporting. Furthermore, existing approaches rarely combine clinically standardized reports with patient-facing explanations and care recommendations. Structured reporting improves consistency and facilitates standardized evaluation \cite{delbrouck2025srrg}, while effective patient communication requires translating professional findings into understandable explanations and actionable guidance. These limitations highlight the need for an integrated PR-specific framework supported by comprehensive multimodal annotations that combines tooth-level diagnosis, localized visual evidence grounding, and standardized clinical reporting with patient-friendly explanations and appropriate care recommendations, while incorporating safeguards against unreliable or unsupported outputs and independent clinical evaluation.

To address these challenges, we propose PeriaTrace, an integrated three-stage deep-learning framework for automated PR analysis and report generation in endodontic practice. Through a single forward pass, its visual network performs anatomical segmentation, tooth localization,  and lesion detection, providing spatial evidence for subsequent tooth-level diagnosis and visual grounding. PeriaTrace generates comprehensive clinical reports with traceable diagnostic findings, alongside accessible explanations, follow-up recommendations, and oral care guidance for patients. To address the scarcity of PR datasets, we also introduce MM-PRAD, to our knowledge the first large-scale multimodal dataset dedicated to PR analysis, comprising 10,192 radiographs from 6,672 patients, with pixel-level segmentation, tooth and caries detection annotations, clinician-authored reports, patient-friendly explanations, and care recommendations. Internal evaluations demonstrated strong performance across segmentation, tooth localization and identification, report generation, and evidence grounding, with improvements over existing models in report generation and visual grounding. External clinical evaluations further showed that most generated reports could be retained without complete rewriting, while automated inference was tens of times faster than manual drafting. A survey of 196 participants without medical backgrounds also revealed strong preferences for patient-friendly explanations, care recommendations, and visually grounded reports, particularly the complete format integrating professional diagnoses with patient-oriented information. Together, these findings highlight the potential of PeriaTrace to improve reporting efficiency, enable tooth-level traceability of diagnostic evidence, and facilitate patient-centred communication in endodontic care. A demonstration video illustrating a potential clinical application of PeriaTrace, including its interactive workflow and report visualization, is provided in Supplementary Video 1.

\begin{figure*}[h]
	\centerline{\includegraphics[width=\linewidth]{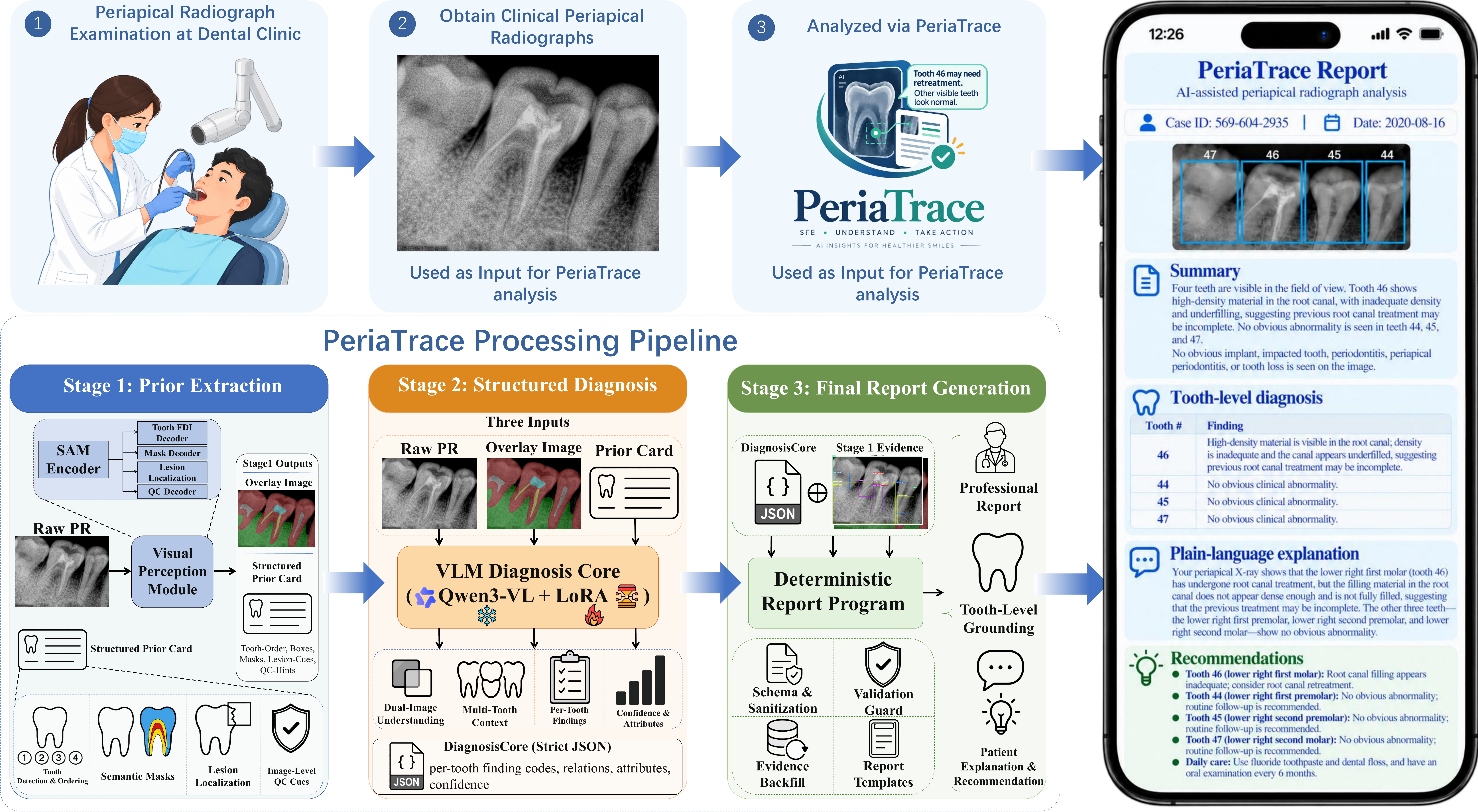}}
	\caption{Clinical workflow and computational pipeline of PeriaTrace. The framework comprises three stages: multitask visual prior extraction, localization-conditioned diagnostic reasoning using a Low-Rank Adaptation (LoRA \cite{hu2021lora})-adapted Qwen3-VL-8B model \cite{bai2025qwen3}, and evidence-grounded report generation. PeriaTrace produces structured tooth-level diagnostic reports with machine-readable visual evidence references, patient-friendly explanations, and care recommendations from periapical radiographs.}
	\label{fig1}
\end{figure*}

\section{Results}\label{Results}
The overall workflow of PeriaTrace is illustrated in Fig. \ref{fig1}. The study included an internal MM-PRAD cohort from the First Affiliated Hospital of Nankai University, comprising 10,192 PRs from 6,672 patients, divided into training, validation, and internal test sets at a ratio of 8:1:1. Two independent external cohorts comprised 1,742 PRs at Tianjin Union Medical Center (TUMC) and 943 PRs at the Second Hospital of Tianjin Medical University (TMUSH), with one radiograph per patient in both cohorts. PeriaTrace was evaluated through quantitative and qualitative analyses of report generation, anatomical segmentation, tooth localization, lesion detection, and visual grounding, complemented by independent assessments from two senior clinicians across the external cohorts and a patient-oriented survey involving individuals without medical backgrounds. These evaluations examined diagnostic accuracy, clinical usability, patient comprehension and preferences, as well as reporting efficiency, providing evidence of the potential of PeriaTrace to support traceable PR analysis and patient-centred clinical reporting.

\begin{figure*}[h]
	\centerline{\includegraphics[width=\linewidth]{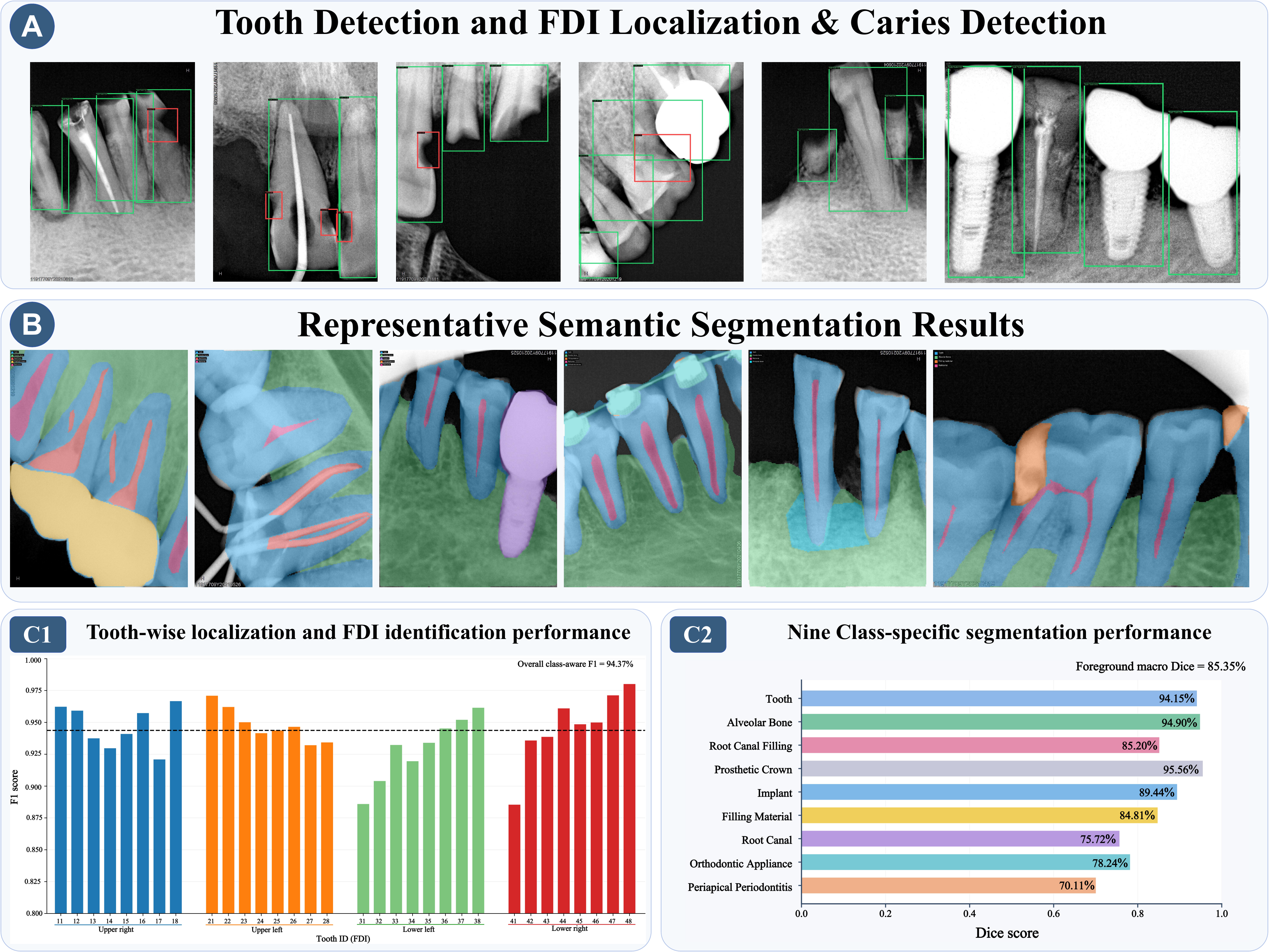}}
	\caption{Qualitative and quantitative performance of PeriaTrace in visual prior extraction. (A) Representative tooth localization and caries detection results. (B) Semantic segmentation results. (C1–C2) Class-specific performance for tooth localization and FDI identification and semantic segmentation, respectively.}
	\label{fig4}
\end{figure*}
\begin{table*}[t]
\centering
\caption{\textbf{Quantitative results on the internal test cohort.}}
\label{tab:quantitative_results}

\setlength{\tabcolsep}{4.2pt}
\renewcommand{\arraystretch}{1.16}

\begin{tabularx}{\textwidth}{
    >{\raggedright\arraybackslash}p{0.27\textwidth}
    >{\raggedright\arraybackslash}X
    >{\centering\arraybackslash}p{0.27\textwidth}
}
\toprule

\rowcolor{PanelGray}
\multicolumn{3}{l}{
\textbf{A. Prior extraction and evidence grounding performance of PeriaTrace}
} \\

\midrule

\rowcolor{HeaderGray}
\textbf{Task}
&
\textbf{Metric}
&
\textbf{Performance}
\\

\midrule

\rowcolor{PriorGray}
QCfor absolute FDI identification
&
F1 / Sensitivity / Specificity
&
74.18\% / 60.36\% / 99.53\%
\\

Tooth localization and identification
&
Localization F1 / Joint localization--FDI F1
&
97.66\% / 94.37\%
\\

\rowcolor{PriorGray}
Caries localization
&
BBox F1 at IoU = 0.3
&
92.17\%
\\

Semantic segmentation
&
Macro Dice / Macro IoU / Pixel accuracy
&
85.35\% / 75.40\% / 93.86\%
\\

\rowcolor{PriorGray}
Report grounding
&
Grounding-tag F1 / Tooth-box / Lesion-bbox / Mask
&
77.93\% / 86.53\% / 71.27\% / 68.43\%
\\

\bottomrule
\end{tabularx}

\vspace{0.9em}

{\scriptsize
\setlength{\tabcolsep}{3.0pt}
\renewcommand{\arraystretch}{1.14}

\begin{tabularx}{\textwidth}{
    >{\raggedright\arraybackslash}p{0.185\textwidth}
    >{\raggedright\arraybackslash}p{0.115\textwidth}
    *{7}{>{\centering\arraybackslash}X}
}
\toprule

\rowcolor{PanelGray}
\multicolumn{9}{l}{
\textbf{B. Comparison of report generation and visual grounding with vision--language baselines}
} \\

\midrule

\rowcolor{HeaderGray}
\textbf{Model}
&
\textbf{Input}
&
\textbf{BLEU-1}
&
\textbf{BLEU-2}
&
\textbf{BLEU-3}
&
\textbf{BLEU-4}
&
\textbf{ROUGE-L}
&
\textbf{CIDEr-lite}
&
\textbf{Grounding F1}
\\

\midrule

\rowcolor{OursGray}
\textbf{PeriaTrace}
&
\textbf{Stage 1 priors}
&
\textbf{0.6870}
&
\textbf{0.6715}
&
\textbf{0.6601}
&
\textbf{0.6504}
&
\textbf{0.6749}
&
\textbf{0.8244}
&
\textbf{77.93\%}
\\

\midrule

Qwen3-VL-8B-Instruct
&
Raw PR
&
0.4462 & 0.3811 & 0.3527 & 0.3343 & 0.4241 & 0.4032 & 5.70\%
\\

\rowcolor{PriorGray}
Qwen3-VL-8B-Instruct
&
+ Stage 1 priors
&
0.5683 & 0.5512 & 0.5394 & 0.5296 & 0.6461 & 0.7515 & 59.71\%
\\

DentVLM
&
Raw PR
&
0.3566 & 0.3563 & 0.3561 & 0.3560 & 0.1883 & 0.2493 & 0.00\%
\\

\rowcolor{PriorGray}
DentVLM
&
+ Stage 1 priors
&
0.5158 & 0.5031 & 0.4946 & 0.4875 & 0.6529 & 0.7534 & 57.96\%
\\

MedMO-8B-Next
&
Raw PR
&
0.2684 & 0.2572 & 0.2515 & 0.2480 & 0.2727 & 0.3350 & 2.02\%
\\

\rowcolor{PriorGray}
MedMO-8B-Next
&
+ Stage 1 priors
&
0.5422 & 0.5236 & 0.5112 & 0.5010 & 0.6349 & 0.7335 & 61.69\%
\\

Lingshu-8B
&
Raw PR
&
0.4146 & 0.3749 & 0.3565 & 0.3445 & 0.3385 & 0.3441 & 3.85\%
\\

\rowcolor{PriorGray}
Lingshu-8B
&
+ Stage 1 priors
&
0.5729 & 0.5534 & 0.5399 & 0.5290 & 0.6490 & 0.7587 & 60.48\%
\\

HuatuoGPT-Vision-7B
&
Raw PR
&
0.2774 & 0.2151 & 0.1927 & 0.1790 & 0.3342 & 0.3967 & 4.97\%
\\

\rowcolor{PriorGray}
HuatuoGPT-Vision-7B
&
+ Stage 1 priors
&
0.5706 & 0.5428 & 0.5240 & 0.5092 & 0.5898 & 0.6864 & 61.00\%
\\

LLaVA-Med-v1.5-Mistral-7B
&
Raw PR
&
0.1882 & 0.1668 & 0.1561 & 0.1493 & 0.3519 & 0.4115 & 3.37\%
\\

\rowcolor{PriorGray}
LLaVA-Med-v1.5-Mistral-7B
&
+ Stage 1 priors
&
0.5298 & 0.5162 & 0.4986 & 0.4807 & 0.6612 & 0.7534 & 62.96\%
\\

\bottomrule
\end{tabularx}
}

\vspace{0.45em}

\begin{minipage}{\textwidth}
\footnotesize
\textit{Notes:}
Segmentation results are macro-averaged over nine foreground anatomical and
pathological classes.
Caries localization was evaluated at an intersection-over-union threshold of 0.3.
Report-generation metrics in Panel B were computed against the
template-standardized ground truth after removal of symbolic grounding tags.
``Raw PR'' denotes evaluation using the original PR, whereas
``+ Stage 1 priors'' additionally provides the visual and structural priors
extracted by Stage 1.
Grounding F1 denotes grounding-tag F1.
\end{minipage}

\end{table*}

\subsection{Quantitative and qualitative analysis}
We evaluated the visual prior extraction performance of PeriaTrace on an internal MM-PRAD cohort of 1,019 adult PRs, with quantitative and qualitative results summarized in Table. \ref{tab:quantitative_results} and Fig. \ref{fig4}. PeriaTrace achieved F1 scores of 97.66\% for tooth localization and 94.37\% for joint localization and FDI identification. Its dedicated quality-control module for unreliable absolute FDI identification achieved an F1 score of 74.18\%, with a sensitivity of 60.36\% and specificity of 99.53\%. When absolute tooth identification was unreliable, the system retained image-relative positional descriptions rather than assigning potentially incorrect FDI numbers. Caries localization achieved an F1 score of 92.17\% at an IoU threshold of 0.3. For semantic segmentation, PeriaTrace achieved a macro Dice of 85.35\%, macro IoU of 75.40\%, and pixel accuracy of 93.86\% across nine foreground classes. Notably, unlike the compared vision--language models, which primarily perform image interpretation and report generation, PeriaTrace integrates tooth localization and identification, caries detection, and pixel-level semantic segmentation into a unified visual analysis framework, providing complementary anatomical and pathological evidence at the tooth, lesion, and pixel levels. Fig. \ref{fig4} further illustrates representative tooth localization and caries detection results (A), semantic segmentation results (B), tooth-wise localization and FDI identification performance (C1), and class-specific segmentation performance across nine foreground classes (C2).

Building on the extracted visual priors, we next evaluated their integration into report generation through the Stage 2 vision–language reasoning module and Stage 3 deterministic reporting pipeline. PeriaTrace was compared with six representative VLMs under two input settings: raw PRs and PRs supplemented with the same Stage 1 priors. Importantly, all models were evaluated using a single shared reporting template and identical terminology and rendering rules, ensuring that equivalent diagnostic findings were expressed in the same standardized format without model-specific templates or wording conventions. This provided a consistent basis for direct comparison across models. Incorporating Stage 1 priors consistently improved report generation across all six baselines, demonstrating the general utility of structured visual evidence across different vision–language architectures.

PeriaTrace achieved BLEU-1, BLEU-2, BLEU-3, and BLEU-4 scores of 0.6870, 0.6715, 0.6601, and 0.6504, respectively, exceeding the best prior-assisted baseline scores of 0.5729, 0.5534, 0.5399, and 0.5296. It also achieved higher ROUGE-L (0.6749 versus 0.6612) and CIDEr-lite (0.8244 versus 0.7587) scores than all baselines. Notably, even DentVLM, specifically developed for dental image understanding, achieved a BLEU-4 score of only 0.4875 with Stage 1 priors, remaining substantially below PeriaTrace. These results demonstrate the advantage of integrating PR-specific visual analysis with vision–language reasoning for comprehensive dental report generation.

The advantage was even more pronounced in visual grounding. Without Stage 1 priors, the six baselines achieved grounding-tag F1 scores of only 0–5.70\%, indicating that they could generate image descriptions but lacked an inherent mechanism to link individual diagnostic statements to specific tooth boxes, lesion regions, or segmentation masks. Providing the same Stage 1 priors increased their grounding F1 scores to 57.96–62.96\%, demonstrating the broad utility of structured visual evidence across different models. Nevertheless, PeriaTrace achieved 77.93\%, outperforming the strongest baseline by 14.97 percentage points. Unlike the baselines, which relied on externally supplied priors to establish these associations, PeriaTrace natively integrates visual prior extraction and evidence binding within its three-stage framework, enabling more reliable correspondence between diagnostic findings and their supporting radiographic evidence.

\begin{figure*}[h]
	\centerline{\includegraphics[width=\linewidth]{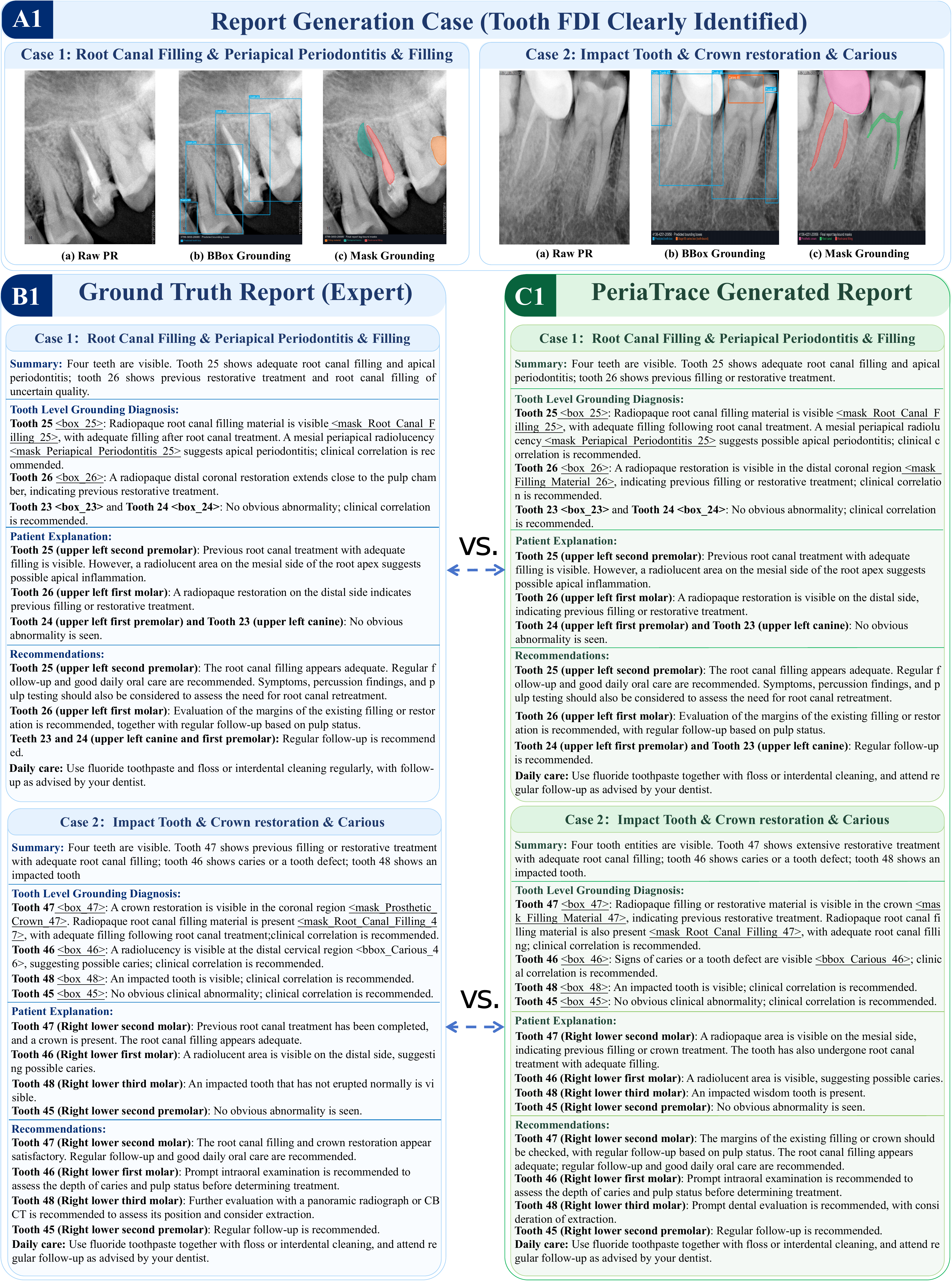}}
	\caption{Representative report generation results with reliable FDI tooth identification. Two cases illustrate raw PRs, bounding-box and mask grounding, and comparisons between expert reference reports and PeriaTrace-generated reports.}
	\label{fig2}
\end{figure*}

\begin{figure*}[h]
	\centerline{\includegraphics[width=\linewidth]{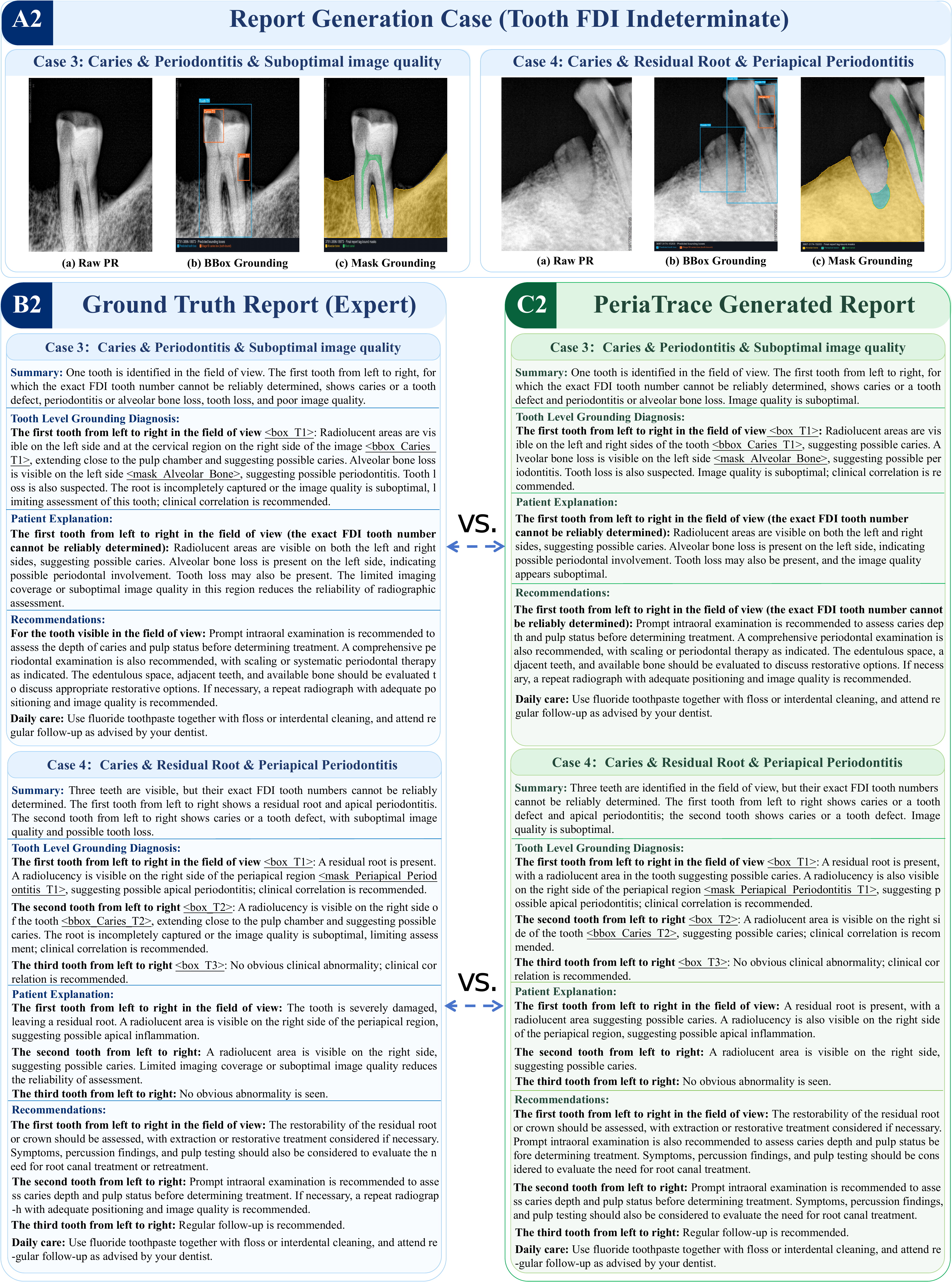}}
	\caption{Representative report generation results with indeterminate FDI tooth identification. Two cases demonstrate relative tooth localization, visual evidence grounding, and comparisons between expert reference reports and PeriaTrace-generated reports.}
	\label{fig3}
\end{figure*}

The qualitative results in Figs. \ref{fig2} and Figs. \ref{fig3} further illustrate the report generation capabilities of PeriaTrace under different tooth identification conditions. When FDI numbers can be reliably determined (Figs. \ref{fig2}), PeriaTrace accurately associates key findings with individual teeth, including root canal filling and periapical periodontitis in tooth 25, caries in tooth 46, and impaction of tooth 48. The corresponding bounding-box and segmentation-mask references establish visual links between diagnostic statements and their supporting image regions. When FDI identification is uncertain (Figs. \ref{fig3}), the system instead adopts relative tooth identifiers (T1–T3), maintaining tooth-level descriptions of caries, alveolar bone loss, residual roots, and periapical abnormalities while incorporating image-quality information. Across these cases, the generated reports show substantial agreement with expert reports in their principal diagnostic findings, patient-friendly explanations, and care recommendations. These examples demonstrate how PeriaTrace integrates visual grounding, tooth identification, and structured reporting to support interpretable diagnostic communication.

\begin{table*}[t]
\centering
\caption{\textbf{External multi-center clinician evaluation of PeriaTrace outputs.}}
\label{tab:external_multicenter}
\setlength{\tabcolsep}{3.0pt}
\renewcommand{\arraystretch}{1.14}
\scriptsize

\begin{tabularx}{\textwidth}{
>{\raggedright\arraybackslash}p{0.285\textwidth}
*{6}{>{\centering\arraybackslash}X}}
\toprule
\rowcolor{PanelGray}
\multicolumn{7}{l}{\textbf{A. Clinical usability of generated reports}}\\
\midrule
\rowcolor{HeaderGray}
\textbf{Outcome} &
\multicolumn{3}{c}{\textbf{TUMC, n=1742}} &
\multicolumn{3}{c}{\textbf{TMUSH, n=943}} \\
\cmidrule(lr){2-4}\cmidrule(lr){5-7}
\rowcolor{HeaderGray}
& \textbf{Reader 1} & \textbf{Reader 2} & \textbf{Both}
& \textbf{Reader 1} & \textbf{Reader 2} & \textbf{Both} \\
\midrule
Image--report concordance, Q1 $\geq$3 (basic/high concordance)
& 1173/1742 (67.34\%) & 1092/1742 (62.69\%) & 1001/1742 (57.46\%)
& 481/943 (51.01\%) & 504/943 (53.45\%) & 440/943 (46.66\%) \\
\rowcolor{LightGray}
Professional report, Q2 A/B (no or minor revision)
& 1081/1742 (62.06\%) & 1113/1742 (63.89\%) & 1054/1742 (60.51\%)
& 451/943 (47.83\%) & 474/943 (50.27\%) & 428/943 (45.39\%) \\
Professional report, Q2 A/B/C (retained without rewrite)
& 1603/1742 (92.02\%) & 1608/1742 (92.31\%) & 1588/1742 (91.16\%)
& 813/943 (86.21\%) & 816/943 (86.53\%) & 797/943 (84.52\%) \\
\rowcolor{LightGray}
Patient explanation, Q3 A/B (direct use or minor revision)
& 1304/1742 (74.86\%) & 1132/1742 (64.98\%) & 1101/1742 (63.20\%)
& 519/943 (55.04\%) & 462/943 (48.99\%) & 436/943 (46.24\%) \\
Care recommendations, Q4 A/B (direct use or minor revision)
& 1324/1742 (76.00\%) & 1195/1742 (68.60\%) & 1122/1742 (64.41\%)
& 526/943 (55.78\%) & 509/943 (53.98\%) & 454/943 (48.14\%) \\
\bottomrule
\end{tabularx}

\vspace{0.65em}

\begin{tabularx}{\textwidth}{
>{\raggedright\arraybackslash}p{0.285\textwidth}
*{6}{>{\centering\arraybackslash}X}}
\toprule
\rowcolor{PanelGray}
\multicolumn{7}{l}{\textbf{B. Grounding evaluation in prespecified subsets}}\\
\midrule
\rowcolor{HeaderGray}
\textbf{Outcome} &
\multicolumn{3}{c}{\textbf{TUMC, n=523}} &
\multicolumn{3}{c}{\textbf{TMUSH, n=283}} \\
\cmidrule(lr){2-4}\cmidrule(lr){5-7}
\rowcolor{HeaderGray}
& \textbf{Reader 1} & \textbf{Reader 2} & \textbf{Both}
& \textbf{Reader 1} & \textbf{Reader 2} & \textbf{Both} \\
\midrule
Grounding helpfulness, G1 $\geq$2 (somewhat/clearly helpful)
& \shortstack[c]{473/523\\[-1pt](90.44\%)}
& \shortstack[c]{443/523\\[-1pt](84.70\%)}
& \shortstack[c]{441/523\\[-1pt](84.32\%)}
& \shortstack[c]{255/283\\[-1pt](90.11\%)}
& \shortstack[c]{236/283\\[-1pt](83.39\%)}
& \shortstack[c]{236/283\\[-1pt](83.39\%)} \\
\rowcolor{LightGray}
Potentially misleading correspondence, G2 = 2 (lower is better)
& \shortstack[c]{57/523\\[-1pt](10.90\%)}
& \shortstack[c]{102/523\\[-1pt](19.50\%)}
& \shortstack[c]{47/523\\[-1pt](8.99\%)}
& \shortstack[c]{31/283\\[-1pt](10.95\%)}
& \shortstack[c]{57/283\\[-1pt](20.14\%)}
& \shortstack[c]{21/283\\[-1pt](7.42\%)} \\
\bottomrule
\end{tabularx}

\vspace{0.65em}

\begin{tabularx}{\textwidth}{
>{\raggedright\arraybackslash}p{0.36\textwidth}
>{\centering\arraybackslash}X
>{\centering\arraybackslash}X
>{\centering\arraybackslash}X
>{\centering\arraybackslash}X}
\toprule
\rowcolor{PanelGray}
\multicolumn{5}{l}{\textbf{C. Inter-reader agreement}}\\
\midrule
\rowcolor{HeaderGray}
\textbf{Assessment} &
\multicolumn{2}{c}{\textbf{TUMC}} &
\multicolumn{2}{c}{\textbf{TMUSH}} \\
\cmidrule(lr){2-3}\cmidrule(lr){4-5}
\rowcolor{HeaderGray}
& \textbf{Exact agreement} & \textbf{Weighted $\kappa$}
& \textbf{Exact agreement} & \textbf{Weighted $\kappa$} \\
\midrule
Q1 Image--report concordance & 78.42\% & 0.786 & 78.05\% & 0.795 \\
\rowcolor{LightGray}
Q2 Professional report usability & 79.56\% & 0.804 & 77.52\% & 0.813 \\
Q3 Patient explanation & 77.67\% & 0.788 & 76.25\% & 0.787 \\
\rowcolor{LightGray}
Q4 Care recommendations & 78.87\% & 0.801 & 77.62\% & 0.800 \\
G1 Grounding helpfulness & 88.15\% & 0.725 & 85.16\% & 0.602 \\
\rowcolor{LightGray}
G2 Grounding deviation & 77.63\% & 0.565 & 71.73\% & 0.345 \\
\bottomrule
\end{tabularx}

\vspace{0.35em}
\begin{minipage}{\textwidth}
\footnotesize
\textit{Notes:}
Q1 was rated on a four-level ordinal scale, with Q1 $\geq$3 indicating basic or high image--report concordance.
For Q2--Q4, A indicates direct use/no revision, B minor revision, C major revision, and D rewrite/not recommended;
A/B therefore denotes at most minor revision, whereas A/B/C denotes that the output could be retained and revised rather than rewritten.
G1 $\geq$2 indicates that the displayed bounding box or mask was at least somewhat helpful for localization.
G2=2 indicates potentially misleading image--text correspondence and is therefore an adverse endpoint.
``Both'' denotes the intersection of cases satisfying the stated criterion for both readers, not an average of their scores.
Exact agreement and linearly weighted Cohen's $\kappa$ were calculated from the paired ordinal ratings before any adjudication.
All paired ratings differed by no more than one adjacent category for Q1--Q4 and G1/G2 in both centers.
Grounding ratings are clinician assessments of localization aids and should not be interpreted as Dice, IoU, or detection accuracy.
\end{minipage}
\end{table*}

\subsection{External Multicentre Validation and Clinical Usability}
To assess the clinical interpretability and usability of PeriaTrace beyond the development centre, we conducted a multicentre evaluation using external PRs from Tianjin Union Medical Center (TUMC; archival database; 1,742 radiographs) and the Second Hospital of Tianjin Medical University (TMUSH; 943 radiographs). The same two dentists, each with more than ten years of clinical experience, assessed the generated outputs against the original radiographs using prespecified criteria, consistent with clinician-based evaluations in previous automated radiology report generation studies \cite{yu2023,hong2025}. Four dimensions were assessed for each case: image–report concordance on a four-point ordinal scale, revision requirements for professional reports, clinical suitability of patient-facing explanations, and appropriateness of care recommendations. The latter three were rated using four-category scales ranging from direct usability to major revision, complete rewriting, or rejection, as appropriate. Visual grounding was additionally assessed in prespecified subsets of approximately 30\% of cases at each centre (523 and 283 radiographs, respectively), examining whether bounding boxes or segmentation masks facilitated the localization of reported findings and whether misleading associations between visual evidence and textual descriptions were present. Reviewer ratings were retained separately rather than averaged. Inter-reader agreement was quantified using exact agreement, agreement within one adjacent category, and linearly weighted Cohen's $\kappa$ \cite{cohen1968}, following established recommendations for reliability and agreement studies \cite{kottner2011}. Centre-specific results are reported in Table. \ref{tab:external_multicenter}. This evaluation characterized the clinical usability and acceptability of PeriaTrace outputs from practising dentists' perspectives, complementing the subsequent patient-centred survey to assess the system's practical utility from both clinical and patient perspectives.

As shown in Table. \ref{tab:external_multicenter}, the external clinical evaluation demonstrated the clinical applicability and usability of PeriaTrace across two independent centres. For image–report concordance (Q1 $\geq$ 3), indicating at least basic agreement with the original radiographs, the two reviewers rated 67.34\% and 62.69\% of TUMC reports and 51.01\% and 53.45\% of TMUSH reports positively, with both reviewers meeting this criterion in 57.46\% and 46.66\% of cases, respectively. Regarding professional report usability, 62.06\% and 63.89\% of reports at TUMC and 47.83\% and 50.27\% at TMUSH required no more than minor revisions (Q2 A/B), with both reviewers reaching this assessment in 60.51\% and 45.39\% of cases, respectively. When reports requiring major revisions but remaining suitable for retention were also included (Q2 A/B/C), both reviewers considered 91.16\% of TUMC reports and 84.52\% of TMUSH reports suitable for retention without complete rewriting. These findings demonstrate that PeriaTrace consistently produces clinically relevant diagnostic reports across independent external cohorts, supporting the generalizability of its three-stage architecture and its potential to provide structured, editable drafts for clinical review and refinement.

We further evaluated the clinical suitability of the patient-facing components. At TUMC, the two senior endodontists rated 74.86\% and 64.98\% of patient-facing explanations, respectively, as directly usable or requiring only minor revisions, with both reviewers assigning this rating to 63.20\% of cases. Care recommendations showed a similar pattern, with corresponding proportions of 76.00\%, 68.60\% and 64.41\%, respectively. At TMUSH, the proportions were lower, following a trend similar to that observed for professional reports. Specifically, 55.04\% and 48.99\% of patient-facing explanations were considered directly usable or suitable for minor revision by the two reviewers, respectively, with both reviewers agreeing on this assessment in 46.24\% of cases. For care recommendations, the corresponding proportions were 55.78\%, 53.98\% and 48.14\%, respectively. Importantly, patient-facing explanations and care recommendations were evaluated as independent outputs, separately from the professional reports, allowing their clinical suitability to be assessed on their own merits. Overall, these findings demonstrate the clinical applicability of PeriaTrace-generated patient-facing explanations and care recommendations, with a proportion considered suitable for direct use or minor revision by clinical experts.

We also assessed the clinical utility of visual grounding using randomly selected subsets from the two centres at the same sampling proportion. As shown in Section B of Table. \ref{tab:external_multicenter}, the visual grounding outputs demonstrated consistent clinical usefulness across both external centres. Within the prespecified evaluation subsets, the two senior endodontists considered the displayed localization cues helpful to at least some extent (G1 $\geq$ 2) in 90.44\% and 84.70\% of TUMC cases, and 90.11\% and 83.39\% of TMUSH cases, respectively. Both reviewers assigned G1 $\geq$ 2 to 84.32\% of cases at TUMC and 83.39\% at TMUSH. We also evaluated potentially misleading image–text correspondence (G2 = 2), for which lower proportions indicate better performance. The two reviewers identified such correspondence in 10.90\% and 19.50\% of TUMC cases, respectively, and in 10.95\% and 20.14\% of TMUSH cases. Collectively, these findings demonstrate that PeriaTrace provides clinically useful visual localization cues across different healthcare settings, supporting the localization and verification of reported findings and highlighting the potential of its visual grounding and prior extraction components for clinical application.

Inter-reader agreement across the six evaluation dimensions is summarized in Section C of Table. \ref{tab:external_multicenter}. Across the four principal clinical assessments (Q1–Q4), exact agreement ranged from 77.67\% to 79.56\% at TUMC and from 76.25\% to 78.05\% at TMUSH, with weighted Cohen's $\kappa$ values of 0.786–0.804 and 0.787–0.813, respectively. Professional report usability (Q2) showed the highest agreement at both centres ($\kappa=0.804$ and $0.813$). For visual grounding, agreement on localization helpfulness (G1) reached 88.15\% and 85.16\%, with corresponding $\kappa$ values of 0.725 and 0.602, whereas agreement on potentially misleading image–text correspondence (G2) was 77.63\% and 71.73\%, with $\kappa$ values of 0.565 and 0.345, respectively. This pattern indicates greater consistency in recognizing the usefulness of localization cues than in identifying potentially misleading associations, reflecting two complementary aspects of visual grounding: facilitating verification of reported findings and ensuring reliable correspondence between visual evidence and diagnostic statements. The consistently high agreement across Q1–Q4, particularly for Q2, further indicates that both clinicians shared similar assessments of report relevance and usability. Together with the usability results in Section A, these findings support the use of PeriaTrace outputs as practical drafts for clinical review and refinement, while the G1 agreement further reinforces the value of visual localization in verifying diagnostic findings and the importance of accurate evidence–text associations.

To further assess workflow efficiency, the two clinicians evaluated approximately 10 representative cases spanning simple, intermediate, and complex presentations after completing the external assessment. PeriaTrace required an average of only 4.41 s per radiograph for automated inference, with some generated reports being directly suitable for use without further modification. For cases requiring clinician review, finalization typically took 1–5 min for straightforward cases and approximately 5–8 min for more complex cases. By comparison, preparing an equivalent report entirely from scratch, including tooth identification, lesion annotation, bounding-box marking, segmentation, professional diagnostic descriptions, patient-friendly explanations, and care recommendations, was estimated by the clinicians to require approximately 15–20 min per case. PeriaTrace substantially improves reporting efficiency, achieving speedups of several tens of times compared with manual report preparation and potentially exceeding a hundredfold when generated reports can be used directly without further revision. By automating the time-consuming image analysis, annotation, and report drafting processes, PeriaTrace allows clinicians to focus primarily on reviewing and refining the generated outputs.

\begin{table*}[t]
\centering
\caption{\textbf{Patient-centered evaluation of report comprehension, actionability, visual grounding, and presentation preferences in 196 non-medical participants.}}
\label{tab:patient_centered_eval}

\setlength{\tabcolsep}{3.2pt}
\renewcommand{\arraystretch}{1.13}
\scriptsize

\begin{tabularx}{\textwidth}{
    >{\centering\arraybackslash}p{0.055\textwidth}
    >{\raggedright\arraybackslash}X
    >{\centering\arraybackslash}p{0.172\textwidth}
    >{\centering\arraybackslash}p{0.172\textwidth}
    >{\centering\arraybackslash}p{0.165\textwidth}
    >{\centering\arraybackslash}p{0.085\textwidth}
}
\toprule
\rowcolor{PanelGray}
\multicolumn{6}{l}{\textbf{A. Professional versus patient-friendly reports}}\\
\midrule
\rowcolor{HeaderGray}
\textbf{Case} & \textbf{Outcome} & \textbf{Patient-friendly} & \textbf{Professional} & \textbf{Effect size} & \textbf{Holm $P$}\\
\midrule

\rowcolor{LightGray}
1 & Objective disease comprehension, n/N (\%) &
79/87 (90.8) & 45/109 (41.3) &
OR 14.04 (6.18--31.92) & $<0.001$ \\

1 & Objective actionability, n/N (\%) &
81/87 (93.1) & 42/109 (38.5) &
OR 21.54 (8.63--53.75) & $<0.001$ \\

\rowcolor{LightGray}
1 & Perceived understandability, mean $\pm$ SD [Top-2, \%] &
4.02 $\pm$ 1.15 [74.7] & 3.21 $\pm$ 1.25 [47.7] &
$d=0.67$ & $<0.001$ \\

1 & Perceived action clarity, mean $\pm$ SD [Top-2, \%] &
3.94 $\pm$ 1.15 [78.2] & 2.71 $\pm$ 1.22 [23.9] &
$d=1.04$ & $<0.001$ \\

\rowcolor{LightGray}
1 & Communication helpfulness, mean $\pm$ SD [Top-2, \%] &
4.09 $\pm$ 1.05 [82.8] & 3.03 $\pm$ 1.37 [45.0] &
$d=0.86$ & $<0.001$ \\
\addlinespace[2pt]

2 & Objective disease comprehension, n/N (\%) &
78/109 (71.56) & 29/87 (33.3) &
OR 5.03 (2.73--9.26) & $<0.001$ \\

\rowcolor{LightGray}
2 & Objective actionability, n/N (\%) &
75/109 (68.81) & 34/87 (39.1) &
OR 3.44 (1.90--6.21) & $<0.001$ \\

2 & Perceived understandability, mean $\pm$ SD [Top-2, \%] &
3.98 $\pm$ 1.16 [76.1] & 2.48 $\pm$ 1.29 [23.0] &
$d=1.23$ & $<0.001$ \\

\rowcolor{LightGray}
2 & Perceived action clarity, mean $\pm$ SD [Top-2, \%] &
3.84 $\pm$ 0.99 [67.9] & 2.40 $\pm$ 1.38 [23.0] &
$d=1.22$ & $<0.001$ \\

2 & Communication helpfulness, mean $\pm$ SD [Top-2, \%] &
4.11 $\pm$ 0.95 [78.9] & 2.62 $\pm$ 1.37 [26.4] &
$d=1.29$ & $<0.001$ \\

\bottomrule
\end{tabularx}

\vspace{0.55em}

\begin{tabularx}{\textwidth}{
    >{\centering\arraybackslash}p{0.055\textwidth}
    >{\raggedright\arraybackslash}X
    >{\centering\arraybackslash}p{0.172\textwidth}
    >{\centering\arraybackslash}p{0.172\textwidth}
    >{\centering\arraybackslash}p{0.165\textwidth}
    >{\centering\arraybackslash}p{0.085\textwidth}
}
\toprule
\rowcolor{PanelGray}
\multicolumn{6}{l}{\textbf{B. BBox versus Mask visual grounding}}\\
\midrule
\rowcolor{HeaderGray}
\textbf{Case} & \textbf{Outcome} & \textbf{Mask} & \textbf{BBox} & \textbf{Effect size} & \textbf{Holm $P$}\\
\midrule

\rowcolor{LightGray}
3 & Objective localization accuracy, n/N (\%) &
74/87 (85.1) & 86/109 (78.9) &
OR 1.52 (0.72--3.22) & 0.707 \\

3 & Ease of localization, mean $\pm$ SD [Top-2, \%] &
4.07 $\pm$ 1.12 [81.6] & 3.91 $\pm$ 1.19 [72.5] &
$d=0.14$ & 0.319 \\

\rowcolor{LightGray}
3 & Image--text correspondence helpfulness, mean $\pm$ SD [Top-2, \%] &
4.18 $\pm$ 1.03 [83.9] & 3.87 $\pm$ 1.28 [71.6] &
$d=0.27$ & 0.226 \\
\addlinespace[2pt]

4 & Objective localization accuracy, n/N (\%) &
72/109 (66.1) & 52/87 (59.8) &
OR 1.31 (0.73--2.35) & 0.707 \\

\rowcolor{LightGray}
4 & Ease of localization, mean $\pm$ SD [Top-2, \%] &
3.88 $\pm$ 1.09 [72.5] & 4.08 $\pm$ 1.05 [80.5] &
$d=-0.19$ & 0.273 \\

4 & Image--text correspondence helpfulness, mean $\pm$ SD [Top-2, \%] &
3.92 $\pm$ 1.02 [74.3] & 3.97 $\pm$ 1.16 [72.4] &
$d=-0.04$ & 0.421 \\

\bottomrule
\end{tabularx}

\vspace{0.55em}

\begin{tabularx}{\textwidth}{
    >{\raggedright\arraybackslash}X
    >{\centering\arraybackslash}p{0.17\textwidth}
    >{\centering\arraybackslash}p{0.24\textwidth}
}
\toprule
\rowcolor{PanelGray}
\multicolumn{3}{l}{\textbf{C. Overall acceptance and presentation preferences}}\\
\midrule
\rowcolor{HeaderGray}
\textbf{Outcome} & \textbf{Result} & \textbf{95\% CI / note}\\
\midrule

Would like future reports to include plain-language explanations and care recommendations &
165/196 (84.2\%) & 78.4--88.6\% \\

\rowcolor{LightGray}
Visual localization cues perceived as helpful &
166/196 (84.7\%) & 79.0--89.1\% \\

Selected a final report format containing a patient-friendly explanation/advice component &
189/196 (96.4\%) & 92.8--98.3\% \\

\rowcolor{LightGray}
\textbf{Preferred the full layered report: professional report + patient-friendly explanation/advice + image localization} &
\textbf{167/196 (85.2\%)} & \textbf{79.6--89.5\%} \\

Direct grounding-format preference &
BBox 26.0\%; Mask 28.1\%; Combined 45.9\% &
Combined was the most common single choice \\

\bottomrule
\end{tabularx}

\vspace{0.35em}
\begin{minipage}{\textwidth}
\footnotesize
\textit{Notes:}
The final analysis included 196 non-medical participants assigned to two counterbalanced survey paths (109 and 87 participants).
For objective outcomes, conditions were compared within each case using Fisher's exact test; odds ratios and 95\% confidence intervals are reported, and Holm correction was applied across the prespecified objective comparisons.
Likert-type outcomes were compared within each case using two-sided Mann--Whitney U tests, with Cohen's $d$ shown as an effect-size summary and Holm-adjusted $P$ values reported.
Top-2 denotes the proportion selecting ``agree'' or ``strongly agree''.
Pooled cross-case percentages are descriptive only because the participant-level within-subject correlation cannot be reconstructed from the aggregate PDF.
BBox, bounding box; CI, confidence interval; OR, odds ratio; SD, standard deviation.
\end{minipage}

\end{table*}

\subsection{Patient-centered evaluation of report comprehension and presentation}
Radiology reports are primarily intended for clinician-to-clinician communication, and their technical terminology may hinder patients' understanding of imaging findings and recommended care \cite{vitzthum2020,cook2017}. Our questionnaire was informed by the Patient Education Materials Assessment Tool (PEMAT), particularly its distinction between understandability and actionability \cite{shoemaker2014}, previous scenario-based studies of non-medical readers' interpretation of radiology reports \cite{chen2026,cho2020}, and patient-centred reporting research using five-point Likert scales to assess perceived understandability, actionability, and communication helpfulness \cite{barrett2021}. Visual grounding was additionally assessed using bounding boxes (BBox) and segmentation masks to link reported findings to image regions, drawing on research into interactive radiology reports, lay-language annotations, and cross-referenced reporting \cite{short2017,cook2017,armstrong2026}. We conducted a web-based, counterbalanced within-participant evaluation involving four fixed PR cases. After informed consent, participants without medical backgrounds assessed professional reports, patient-friendly explanations with care recommendations, and BBox- and mask-based visual presentations across different cases. Two counterbalanced survey paths controlled the assignment of report formats and grounding methods to reduce case-specific and presentation-order effects. Multiple-choice questions evaluated comprehension of principal imaging findings, appropriate follow-up actions, and abnormality localization, while five-point Likert items assessed perceived understandability, actionability, communication helpfulness, and image–text correspondence. Overall preferences for report formats and visual grounding methods were collected after all case-level assessments.

The four cases were selected to assess two complementary aspects of patient understanding. Cases 1 and 2 evaluated comprehension of diagnostic findings and recommended clinical actions. Case 1 showed caries or tooth structure loss with periapical inflammation in the mandibular left second premolar (tooth 35), with no evident abnormalities in other visible teeth. Case 2 presented caries or tooth structure loss in the mandibular left first premolar (tooth 34), alongside extensive restoration and satisfactory root canal filling in the mandibular left second premolar (tooth 35). Participants were required to identify the principal findings and appropriate follow-up actions. Cases 3 and 4 assessed radiographic evidence localization. Case 3 showed a previously root canal-treated mandibular left canine (tooth 33) with signs of periapical inflammation, requiring identification of the affected tooth. Case 4 featured radiopaque structures resembling implant or orthodontic anchorage screws near teeth 35 and 36, requiring localization of these treatment-related findings. Together, these cases assessed comprehension of diagnostic findings and care recommendations, as well as the correspondence between textual descriptions and radiographic evidence. Quantitative results are presented in Table. \ref{tab:patient_centered_eval}, and the complete Chinese and English versions of the patient-centered questionnaire are provided in Supplementary PDFs 1 and 2, respectively.

As shown in Table. \ref{tab:patient_centered_eval}, patient-friendly reports achieved better objective performance and more favourable subjective assessments than professional reports among 196 non-medical participants. In Case 1, the patient-friendly format improved accuracy in identifying principal radiographic findings (90.8\% versus 41.3\%; OR 14.04, 95\% CI 6.18–31.92) and appropriate follow-up actions (93.1\% versus 38.5\%; OR 21.54, 95\% CI 8.63–53.75), both with Holm-adjusted $P<0.001$. Similar improvements were observed in Case 2 for disease comprehension (71.56\% versus 33.3\%; OR 5.03, 95\% CI 2.73–9.26) and follow-up actions (68.81\% versus 39.1\%; OR 3.44, 95\% CI 1.90–6.21), again with Holm-adjusted $P<0.001$. Subjective assessments likewise favoured patient-friendly reports across both cases in perceived understandability, action clarity, and communication helpfulness, with moderate to large effect sizes (Cohen's $d=0.67$–1.29; all Holm-adjusted $P<0.001$). These consistent objective and subjective improvements suggest that patient-friendly reporting offers benefits beyond simplified terminology, helping non-medical readers understand dental findings, identify appropriate care actions, and communicate more effectively with dental professionals. The findings support integrating accessible explanations and care recommendations into radiographic reports.

Visual grounding provides a complementary means of linking report content to the underlying radiographic evidence. As shown in Section B of Table. \ref{tab:patient_centered_eval}, mask-based presentation achieved numerically higher objective localization accuracy than bounding-box (BBox) presentation in both Case 3 (85.1\% versus 78.9\%) and Case 4 (66.1\% versus 59.8\%), although neither difference reached statistical significance (both Holm-adjusted $P=0.707$). Subjective assessments likewise showed no consistent advantage for either presentation method. In Case 3, masks received higher ratings for ease of localization (4.07 versus 3.91) and image–text correspondence helpfulness (4.18 versus 3.87), whereas in Case 4, bounding boxes received slightly higher ratings for ease of localization (4.08 versus 3.88). None of these differences was statistically significant after Holm correction. These findings suggest that bounding boxes and segmentation masks offer complementary forms of visual evidence rather than a single consistently superior presentation method. Bounding boxes provide intuitive object-level or tooth-level localization, whereas segmentation masks enable more detailed region-level correspondence between radiographic findings and textual descriptions. Their complementary characteristics support a flexible visual grounding strategy that accommodates different localization needs and levels of anatomical detail in patient-facing radiographic reports.

Overall acceptance and presentation preferences further supported PeriaTrace's layered reporting strategy (Section C, Table. \ref{tab:patient_centered_eval}). Most participants preferred future reports to include plain-language explanations and care recommendations (84.2\%) and found visual localization cues helpful (84.7\%). Notably, 96.4\% selected formats incorporating patient-friendly explanations and advice, with 85.2\% preferring the complete integration of professional diagnoses, patient-facing explanations, care recommendations, and visual localization, while only 3.1\% preferred professional reports alone. Preferences for bounding boxes (BBox; 26.0\%) and segmentation masks (28.1\%) were comparable, whereas their combined use was most frequently preferred (45.9\%), suggesting their complementary value. These findings indicate that participants favoured preserving professional diagnostic information while incorporating accessible explanations, actionable recommendations, and visually grounding evidence. By addressing the interconnected needs of clinicians and patients within a single report, PeriaTrace's layered design extends radiographic reporting beyond clinical documentation, facilitating patient understanding of dental findings and recommended care, as well as subsequent communication with dental professionals.

\section{Discussion}\label{Discussion}
This study introduces PeriaTrace, an integrated framework for tooth-level diagnosis, evidence-grounded reporting, and patient-centred communication from PRs. Supported by the large-scale, professionally annotated MM-PRAD dataset, PeriaTrace combines anatomical and pathological image analysis with structured diagnostic reasoning to generate clinically oriented reports, grounding visual evidence, and accessible patient explanations and care recommendations. Its performance was assessed through internal quantitative and qualitative experiments, external multicentre clinician evaluations, and a non-medical participant survey. Together, these evaluations highlight the potential of PeriaTrace to bridge automated radiographic interpretation and clinical reporting while addressing the complementary information needs of dental professionals and patients.

PeriaTrace was developed to address the limited applicability of existing medical LLMs and MLLMs to tooth-specific PR interpretation and reporting. Although dental-specialized MLLMs have advanced image understanding and report generation, most have focused on panoramic radiographs and intraoral photographs, leaving PR-specific requirements less explored. PR interpretation demands precise associations between individual teeth, localized abnormalities, treatment-related structures, and anatomical locations, requiring each diagnostic statement to be linked to its corresponding tooth and identifiable radiographic evidence rather than merely describing the overall image. To our knowledge, PeriaTrace is the first end-to-end PR-specific framework integrating tooth-level grounding diagnosis, evidence-grounded professional reports, and patient-facing explanations and care recommendations. By explicitly connecting diagnostic findings with spatial evidence and information communicated to clinicians and patients, it extends conventional dental report generation towards more transparent and clinically meaningful reporting.

Meeting these requirements is complicated by the inherent variability of PR imaging. Each radiograph captures a localized region, with substantial variation in the number, arrangement, and visibility of teeth across examinations. Receptor positioning, projection angle, exposure, and anatomical superimposition further affect image quality and feature visibility \cite{sarsam2025role}. Reliable tooth identification is therefore essential, as incorrect FDI assignments may propagate into subsequent diagnostic statements. To address this variability, PeriaTrace extracts tooth locations and identities, caries bounding boxes, and segmentation masks of anatomical structures and lesions, associating these visual priors with corresponding findings to guide report generation. When absolute FDI identification is unreliable, relative tooth numbering based on the order of visible teeth is adopted instead. Comparisons with six vision–language baselines showed that incorporating Stage 1 priors consistently improved report similarity and grounding performance, although PeriaTrace outperformed all prior-assisted baselines across the evaluated metrics. These results demonstrate the value of explicit visual priors while highlighting the importance of integrating them into diagnostic reasoning and maintaining evidence correspondence throughout report generation, as achieved by PeriaTrace's three-stage architecture.

The comparative evaluation further substantiated the advantages of PeriaTrace over six general-purpose, medical, and dental vision–language baselines. PeriaTrace consistently achieved superior report similarity and visual grounding, with particularly pronounced gains in mask-level evidence association. Notably, the baseline models exhibited negligible grounding capability when provided with raw PRs, indicating that explicitly linking diagnostic statements to localized radiographic evidence is not an inherent capability of existing VLMs. Providing the same Stage 1 visual priors substantially improved their grounding performance, demonstrating the general utility of structured tooth- and lesion-level evidence across different model architectures. Nevertheless, all prior-assisted baselines remained below PeriaTrace, indicating that visual priors alone are insufficient to achieve the same level of report quality and fine-grained evidence correspondence. This distinction is clinically important because textual similarity alone cannot establish whether a diagnostic statement refers to the correct tooth or radiographic region. Although existing dental MLLMs can generate global reports and support basic visual question answering, their ability to associate individual diagnostic statements with fine-grained visual evidence remains limited. By maintaining these associations throughout report generation, PeriaTrace produces visually traceable grounding diagnostic records that facilitate clinical verification, particularly in endodontic practice, where abnormalities and treatment-related structures in adjacent teeth require precise differentiation.

External clinician assessments and the patient-centred survey provided complementary evidence of PeriaTrace's practical value. Across TUMC and TMUSH, both reviewers considered 91.16\% and 84.52\% of professional reports, respectively, suitable for retention and revision without complete rewriting. Visual localization cues were also considered at least somewhat helpful in 84.32\% and 83.39\% of the assessed cases, respectively. The time assessment further demonstrated a several-tens-fold reduction in report generation time compared with comprehensive manual drafting. More importantly, by shifting clinicians' workload from constructing reports from scratch to verifying and refining structured outputs, PeriaTrace may make comprehensive, evidence-linked reporting more feasible in routine practice while retaining professional oversight. From the patient perspective, 84.2\% of the 196 non-medical participants favoured including plain-language explanations and care recommendations in future radiographic reports, 84.7\% found visual localization helpful, and 85.2\% preferred the fully layered presentation. Together, these findings support the practical value of combining efficient, clinically usable reports with accessible explanations and traceable grounding visual evidence. By accommodating the complementary needs of clinicians and patients within a unified reporting framework, PeriaTrace extends AI-assisted dental reporting beyond diagnostic text generation towards more effective clinical review and patient-centred communication.

This study has several limitations. First, all cohorts were collected within the same geographical region, and variations between external cohorts suggest that imaging equipment, acquisition protocols, patient populations, and clinical practices may affect generalizability. Prospective validation across more diverse institutions and regions is therefore warranted. Second, errors in upstream tooth localization, identification, lesion detection, and segmentation may propagate through the staged architecture to subsequent diagnostic descriptions and visual grounding. Uncertainty quantification, open-set recognition, and confidence-aware processing may further improve robustness, particularly for rare or previously unseen abnormalities. Third, the patient-centred evaluation relied on predefined cases and a web-based questionnaire rather than a prospective clinical workflow; the effects on clinical efficiency, diagnostic decision-making, patient behaviour, and clinical outcomes remain to be established in real-world studies. Finally, as an AI-assisted tool, PeriaTrace may produce inaccurate or incomplete findings and is intended to support rather than replace clinical judgement. All generated outputs should therefore be reviewed by qualified dental professionals before informing clinical decisions or patient care.

\section{Methods}\label{Methods}
\subsection{Ethics}
This study complied with the Declaration of Helsinki and was approved by the Ethics Committees of the First Affiliated Hospital of Nankai University (Tianjin People's Hospital; approval no. 2025-B114) and the Second Hospital of Tianjin Medical University (approval no. KY2026K432). Questionnaire participants were informed of the study purpose, procedures, and voluntary nature of participation, and provided informed consent before completion. Only responses from consenting participants were analyzed.

\subsection{Data Acquisition}
The study included internal and external cohorts, with experienced radiologists screening radiographs to exclude those unsuitable for clinical interpretation. The internal MM-PRAD dataset comprised 10,192 PRs from 6,672 patients, acquired between 2022 and 2024 at the First Affiliated Hospital of Nankai University using a Runyes RAY68 (W) X-ray unit and Handy HDR-500A sensor. Building on our previous datasets \cite{zhou2025prad,pard++}, MM-PRAD underwent further image review and replacement. The original nine-class segmentation scheme from PRAD (Tooth, Alveolar Bone, Pulp, Root Canal Filling, Denture Crown, Dental Fillings, Implant, Orthodontic Devices, and Apical Periodontitis) was retained, with existing masks reviewed and corrected and newly included images receiving corresponding pixel-level annotations. Building on these refined segmentation masks, extensive additional annotations were performed, including structured professional reports, patient-friendly explanations, care recommendations, tooth bounding boxes with FDI or relative numbering, and caries bounding boxes. Four clinicians, comprising two chief physicians with over 20 years of experience and two associate chief physicians with over 10 years, were divided into two groups for initial annotation, with cases evenly allocated and subsequently exchanged for cross-review. Disagreements were resolved through joint discussion, with a senior radiologist providing imaging and reporting consultation and a computer science researcher consolidating annotations into standardized machine-readable formats. The clinicians established and iteratively refined a unified reporting template before its adoption for Stage 3 report construction. Dataset construction took approximately 20 months.

The external evaluation comprised two cohorts. The TUMC cohort included 1,742 PRs from 1,742 distinct patients in the archival database of Tianjin Union Medical Center, whose records predated its reorganization as the First Affiliated Hospital of Nankai University. Acquired between 2018 and 2020 using an eXpert DC dental X-ray system (Gendex Dental Systems) and its associated intraoral image receptor, these radiographs differed from MM-PRAD in imaging equipment, resolution, and acquisition formats. The TMUSH cohort comprised 943 PRs from 943 distinct patients, acquired using an R180-2 dental X-ray unit and an F200 digital intraoral imaging plate system (Fussen Technology Co., Ltd., Shenzhen, China). Both cohorts followed the MM-PRAD selection criteria, excluding images with inadequate projection, insufficient diagnostic quality, or limited clinical relevance. Unlike MM-PRAD, neither cohort included expert-annotated segmentation masks or reference diagnostic reports. Instead, two experienced endodontists independently assessed the generated outputs for image–report concordance, clinical usability, patient-facing content, and visual grounding, enabling external clinical evaluation across different imaging systems and institutional settings.

\begin{figure*}[t]
	\centerline{\includegraphics[width=\linewidth]{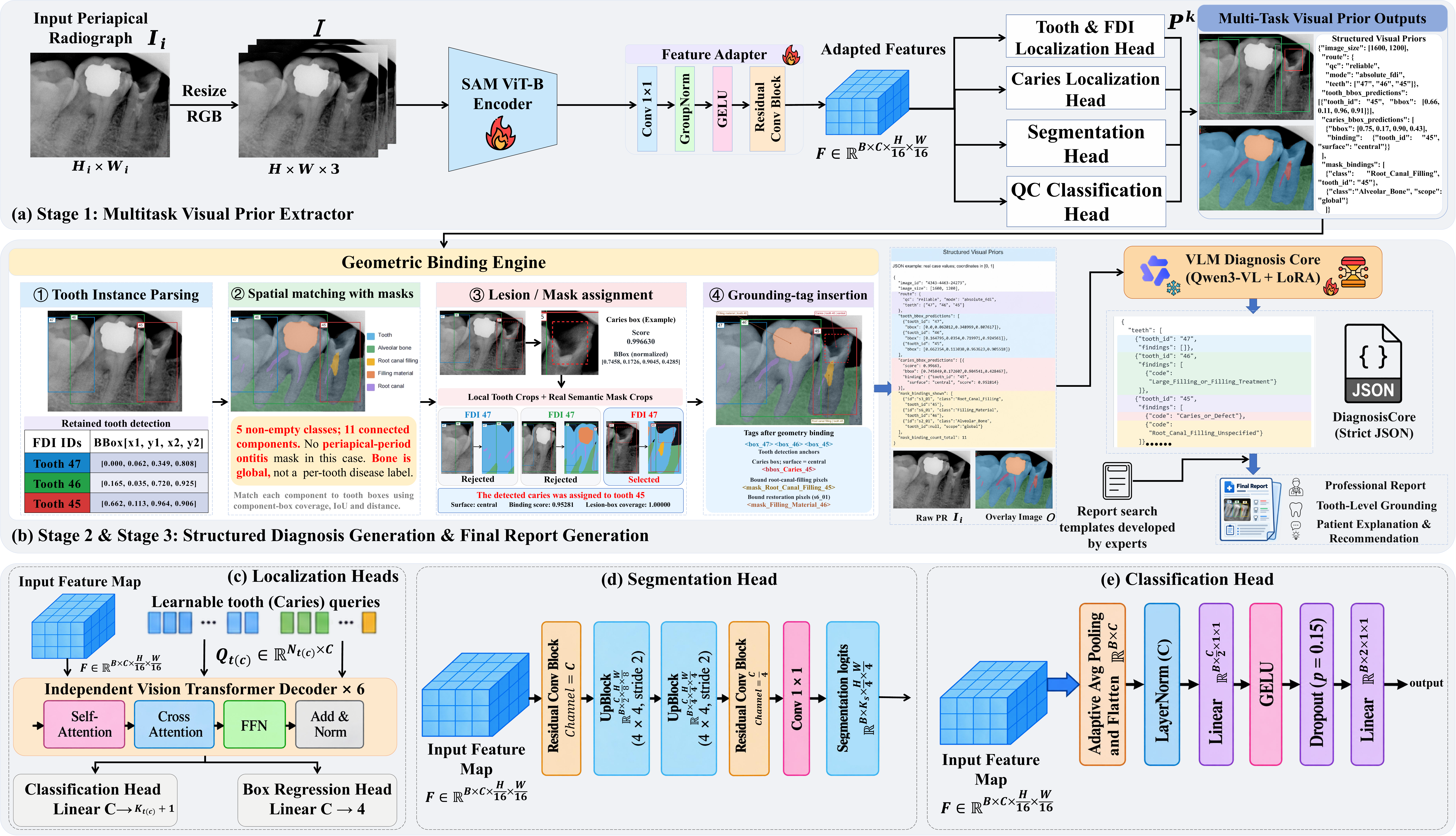}}
	\caption{Detailed architecture and processing workflow of PeriaTrace. (a) Multitask visual prior extraction using a shared SAM ViT-B encoder. (b) Geometric evidence binding, localization-conditioned diagnosis with Qwen3-VL, and evidence-grounded report generation. (c–e) Detailed architectures of the localization, segmentation, and quality-control heads, respectively.}
	\label{fig5}
\end{figure*}

\subsection{PeriaTrace Architecture}
\subsubsection{Overview}
PeriaTrace comprises three stages integrating multitask visual prior extraction, localization-conditioned diagnostic inference, and evidence-grounded report generation, as illustrated in Fig. \ref{fig5}. Given an input PR image $I$, Stage 1 converts it into a three-channel representation and resizes it to $H \times W$. A Segment Anything Model (SAM) \cite{kirillov2023segment} ViT-B image encoder followed by a trainable feature adapter extracts a shared feature map $F$ at one-sixteenth of the input spatial resolution. Four task-specific heads operate on $F$ to generate visual priors $P=\{P^k \mid k\in\mathcal{K}\}$, where $\mathcal{K}=\{\mathrm{tooth},\mathrm{caries},\mathrm{seg},\mathrm{QC}\}$, corresponding to tooth localization and FDI identification, caries localization, semantic segmentation, and QC of absolute FDI identification, respectively. Stage 2 employs a geometric binding engine to perform QC-aware tooth identification, lesion-to-tooth assignment, and segmentation-component association, generating localization priors $L$ and an annotation overlay $O$. The original radiograph $I$, overlay $O$, and localization priors $L$ are then provided to a LoRA-adapted Qwen3-VL-8B model to generate structured tooth-level diagnostic findings $Y$. In Stage 3, deterministic validation and evidence-grounding procedures verify the generated findings against the visual priors $P$, followed by template-based rendering to produce professional diagnostic reports, tooth-level visual references, patient-friendly explanations, and care recommendations, collectively denoted by $Z$. The visual network and vision--language model are optimized separately, without end-to-end gradient propagation through the intermediate geometric binding and report-rendering procedures.

\subsubsection{Multitask Visual Prior Extraction}
Stage 1 employs a shared SAM ViT-B encoder with four task-specific heads to extract anatomical and spatial priors from PRs (Fig. \ref{fig5}). Each input image $I$ is converted to three channels, resized to $1024\times1024$ using bicubic interpolation, and normalized using SAM channel statistics. A pretrained SAM ViT-B encoder and a trainable feature adapter, comprising a $1\times1$ convolution, group normalization, GELU activation, and a residual convolutional block, produce shared features $F\in\mathbb{R}^{256\times64\times64}$, as formulated in Equation. \ref{eq:stage1_features}.

\begin{equation}
F=\mathcal{A}\left(\mathcal{E}_{\mathrm{SAM}}(\widetilde{I})\right),
\label{eq:stage1_features}
\end{equation}
Here, $\widetilde{I}$ denotes the preprocessed image, while $\mathcal{E}_{\mathrm{SAM}}$ and $\mathcal{A}$ represent the SAM encoder and feature adapter, respectively. Both components are updated during multitask training. Four parallel prediction heads operate on $F$ to perform tooth localization and FDI identification, caries detection, semantic segmentation, and QC classification. The tooth and caries heads use independent DETR-style decoders \cite{carion2020end}, each with six Transformer layers, eight attention heads, and a feed-forward dimension of 1024. They employ 80 and 32 learnable queries, respectively, to predict class probabilities and normalized bounding boxes. The tooth head predicts 34 source categories plus a no-object class, whereas the caries head performs binary lesion detection. The segmentation head uses residual convolutional and upsampling blocks to produce ten-class logits, including background, at $256\times256$ resolution. The QC head applies global average pooling and an MLP to assess the reliability of absolute FDI identification. The task-specific predictions are defined in Equation. \ref{eq:stage1_heads}, where $k$ indexes the four tasks in $\mathcal{K}$, and their outputs collectively constitute the visual prior set $P$.

\begin{equation}
P^k=\operatorname{Head}^{k}(F),\quad k\in\mathcal{K}.
\label{eq:stage1_heads}
\end{equation}

\subsubsection{Multigranular Evidence Integration and Diagnostic Reasoning}
Stage 1 produces complementary visual predictions, including tooth identities, bounding boxes, segmentation masks, and quality assessments, which provide the initial anatomical and spatial information required for diagnostic inference. Stage 2 further organizes these predictions through geometric association into multigranular priors, encompassing image-level context, tooth-level anatomy, and localized pathological or treatment-related features. Rather than directly converting visual predictions into diagnostic conclusions, these priors are presented as contextual evidence with confidence estimates, spatial relationships, and alternative tooth assignments where applicable, enabling the VLM to assess their reliability and clinical relevance. A LoRA-adapted Qwen3-VL-8B serves as the diagnostic reasoning component, integrating these complementary cues to determine which findings are clinically meaningful, which associations warrant acceptance, and which uncertain predictions require further interpretation. For example, the model can assess root canal filling quality by jointly interpreting anatomical and treatment-related features or distinguish carious defects from restoration-related radiographic appearances. The model additionally receives both the original radiograph and an annotation overlay, preserving direct access to radiographic details while explicitly highlighting the spatial correspondence between teeth and localized findings. This combination enables the model to interpret and critically evaluate the structured priors against the underlying image evidence, rather than relying solely on intermediate visual predictions, ultimately producing coherent tooth-level diagnostic representations.

To establish a consistent tooth reference system, the tooth detections $P^{\mathrm{tooth}}$ are first refined through confidence filtering and non-maximum suppression to obtain valid FDI and generic tooth candidates, denoted by $T_{\mathrm{FDI}}$ and $T_{\mathrm{gen}}$, respectively. The QC prediction $P^{\mathrm{QC}}$ then guides tooth identification through the routing mechanism defined in Equation. \ref{eq:stage2_routing}. Here, $\rho$ denotes the predicted probability of unreliable absolute FDI identification, and $\tau_{\mathrm{QC}}=0.85$ is the QC threshold. Images with unreliable FDI identification are assigned to the relative mode, in which detected teeth are numbered sequentially from left to right, while reliable detections retain their absolute FDI identifiers. The resulting tooth set $T$ provides a unified spatial reference for subsequent lesion and anatomical evidence association.

\begin{equation}
\begin{aligned}
m &=
\begin{cases}
\mathrm{relative},
& \rho \geq \tau_{\mathrm{QC}}
\lor (T_{\mathrm{FDI}}=\varnothing \land T_{\mathrm{gen}}\neq\varnothing),\\
\mathrm{absolute\_fdi},
& \mathrm{otherwise},
\end{cases}\\
T &=
\begin{cases}
\{\mathrm{T1},\ldots,\mathrm{Tn}\},
& m=\mathrm{relative},\\
T_{\mathrm{FDI}},
& m=\mathrm{absolute\_fdi}.
\end{cases}
\end{aligned}
\label{eq:stage2_routing}
\end{equation}

The remaining visual priors are organized through preliminary geometric association to provide tooth-level localization cues for subsequent diagnostic reasoning. For each caries candidate $\ell$ from $P^{\mathrm{caries}}$, a weighted score evaluates its potential association with each detected tooth by combining spatial overlap, positional containment, centroid proximity, and tooth detection confidence, as defined in Equation. \ref{eq:stage2_caries_binding}. Here, $u_{\ell t}$ represents lesion-tooth overlap, $e_{\ell t}$ indicates positional containment, $v_{\ell t}$ measures normalized centroid proximity, and $p_t$ denotes tooth detection confidence. The highest-scoring association is retained as a preliminary candidate when its score exceeds the threshold $\tau_b=0.45$, providing an initial spatial reference for further interpretation.

\begin{equation}
\begin{aligned}
s(\ell,t)&=0.40u_{\ell t}+0.30e_{\ell t}
+0.20v_{\ell t}+0.10p_t,\\
\hat{t}(\ell)&=
\begin{cases}
\displaystyle\arg\max_{t\in T}s(\ell,t),
& \max_{t\in T}s(\ell,t)\geq\tau_b,\\
\varnothing,&\mathrm{otherwise}.
\end{cases}
\end{aligned}
\label{eq:stage2_caries_binding}
\end{equation}

Similarly, the segmentation predictions $P^{\mathrm{seg}}$ are decomposed into eight-connected components, each representing a candidate anatomical, treatment-related, or pathological region. As formulated in Equation. \ref{eq:stage2_mask_binding}, local components are provisionally associated with individual teeth through lexicographic ranking (($\mathrm{lex}$)) of component coverage, bounding-box intersection over union (IoU), and normalized centroid distance, prioritized in that order. Here, $b_j$ and $b_t$ denote the component and tooth bounding boxes, respectively, while $d(j,t)$ represents their normalized centroid distance. Extensive alveolar bone components are preserved as global anatomical context. Together, these preliminary associations provide spatially organized, multigranular evidence for Qwen3-VL to interpret alongside the original radiograph during tooth-specific diagnostic reasoning.

\begin{equation}
\hat{t}(j)=
\operatorname*{arg\,max}_{t\in T}^{\mathrm{lex}}
\left(
\frac{|b_j\cap b_t|}{|b_j|},
\operatorname{IoU}(b_j,b_t),
-d(j,t)
\right).
\label{eq:stage2_mask_binding}
\end{equation}

The preliminary geometric associations are further organized into a compact localization prior $L$, incorporating the QC-guided identification mode, tooth locations, candidate lesion associations, and segmentation-derived components with their spatial relationships and confidence levels. In parallel, an annotation overlay $O$ is constructed by superimposing tooth identifiers and anatomical or lesion bounding boxes onto the original radiograph. The original image $I$, overlay $O$, and structured prior $L$ are jointly provided to a LoRA-adapted Qwen3-VL-8B, enabling the model to integrate multigranular contextual cues with direct visual evidence for tooth-specific diagnostic reasoning. Supervised by structured diagnostic targets derived from clinician-reviewed canonical reports, the model learns to interpret anatomical and pathological features, assess the relevance of preliminary associations, and formulate tooth-level diagnostic findings. Its output $Y$ is a structured JSON representation containing tooth identifiers, normality indicators, diagnostic finding codes, and associated attributes. These predictions subsequently undergo evidence-based validation and report rendering in Stage 3.

\subsubsection{Deterministic Validation and Evidence-Grounded Report Generation}

In Stage 3, the structured diagnostic output $Y$ generated by Qwen3-VL-8B undergoes deterministic validation and visual evidence association. The output is first parsed as JSON, with balanced-object extraction applied if direct parsing fails. The extracted predictions are validated against the routed tooth set $T$ and a predefined vocabulary of permissible diagnostic codes, removing invalid tooth identifiers, unsupported codes, and duplicate entries while reconciling normality indicators with retained findings. To mitigate potential tooth-wise duplication, a consistency check is applied when identical diagnostic codes are assigned to all detected teeth, using available localization evidence to identify unsupported repetitive assignments. The retained findings are then linked to corresponding tooth bounding boxes, caries lesion boxes, and segmentation-derived regions through the geometric associations established in Stage 2. The resulting representation preserves tooth-level diagnoses, corresponding visual evidence references, and auxiliary prediction provenance.

The validated representation is then rendered into the final report $Z$ using predefined, clinician-designed templates, comprising professional tooth-level diagnostic descriptions, patient-friendly explanations, and care recommendations. Machine-readable evidence tags are embedded alongside relevant diagnostic statements to reference available tooth bounding boxes, caries lesion boxes, and segmentation regions, enabling the user interface to retrieve and visualize the corresponding radiographic evidence. Tags are generated only for available visual evidence, while all report sections are derived from the same validated findings to maintain consistency between professional diagnoses, patient explanations, and care recommendations.

\subsection{Evaluation metrics}

\begin{table*}[t]
\centering
\caption{\textbf{Definitions of quantitative evaluation metrics and statistical measures.}}
\label{tab:evaluation_metrics}

\setlength{\tabcolsep}{2.5pt}
\setlength{\extrarowheight}{3pt}
\renewcommand{\arraystretch}{1.28}
\scriptsize

\begin{tabular}{|
>{\centering\arraybackslash}m{0.155\textwidth}|
>{\centering\arraybackslash}m{0.315\textwidth}|
>{\centering\arraybackslash}m{0.155\textwidth}|
>{\centering\arraybackslash}m{0.315\textwidth}|}

\hline
\rowcolor{HeaderGray}
\textbf{Metric} & \textbf{Definition} &
\textbf{Metric} & \textbf{Definition} \\
\hline

\rowcolor{PanelGray}
\multicolumn{4}{|l|}{\textbf{A. Detection}} \\
\hline

Bounding-box IoU &
$\displaystyle
\operatorname{IoU}(b_p,b_g)
=\frac{|b_p\cap b_g|}{|b_p\cup b_g|}
$ &
Precision &
$\displaystyle
\mathrm{Precision}
=\frac{\mathrm{TP}}{\mathrm{TP}+\mathrm{FP}}
$ \\
\hline

Recall &
$\displaystyle
\mathrm{Recall}
=\frac{\mathrm{TP}}{\mathrm{TP}+\mathrm{FN}}
$ &
F1 &
$\displaystyle
F_1
=\frac{2\mathrm{TP}}
{2\mathrm{TP}+\mathrm{FP}+\mathrm{FN}}
$ \\
\hline

\rowcolor{PanelGray}
\multicolumn{4}{|l|}{\textbf{B. Semantic segmentation}} \\
\hline

Class-specific Dice &
$\displaystyle
\mathrm{Dice}_k
=\frac{2|A_k\cap G_k|}{|A_k|+|G_k|}
$ &
Class-specific IoU &
$\displaystyle
\mathrm{IoU}_k
=\frac{|A_k\cap G_k|}{|A_k\cup G_k|}
$ \\
\hline

MacroDice &
$\displaystyle
\mathrm{MacroDice}
=\frac{1}{K_s-1}
\sum_{k=1}^{K_s-1}\mathrm{Dice}_k
$ &
MacroIoU &
$\displaystyle
\mathrm{MacroIoU}
=\frac{1}{K_s-1}
\sum_{k=1}^{K_s-1}\mathrm{IoU}_k
$ \\
\hline

Pixel accuracy &
$\displaystyle
\mathrm{PixelAcc}
=\frac{\sum_{k=0}^{K_s-1}n_{kk}}
{\sum_{k=0}^{K_s-1}\sum_{j=0}^{K_s-1}n_{kj}}
$ &
&
\\
\hline

\rowcolor{PanelGray}
\multicolumn{4}{|l|}{\textbf{C. Report generation and evidence grounding}} \\
\hline

BLEU-1 to BLEU-4 &
$\displaystyle
\mathrm{BLEU}_n
=\mathrm{BP}\exp\!\left(
\frac{1}{n}\sum_{k=1}^{n}\log p_k
\right)
$ &
ROUGE-L &
$\displaystyle
\mathrm{ROUGE}\text{-}L
=\frac{(1+\beta^2)P_{\mathrm L}R_{\mathrm L}}
{R_{\mathrm L}+\beta^2P_{\mathrm L}}
$ \\
\hline

CIDEr-lite &
$\displaystyle
\mathrm{CIDEr}_{\mathrm{lite}}(h,r)
=\sum_{n\in\mathcal N}\omega_n
\operatorname{cos}
\bigl(\mathbf v_n(h),\mathbf v_n(r)\bigr)
$ &
Grounding-tag F1 &
$\displaystyle
F_{1,\mathrm{ground}}
=\frac{2M_g}
{N_g^{\mathrm{pred}}+N_g^{\mathrm{ref}}}
$ \\
\hline

\rowcolor{PanelGray}
\multicolumn{4}{|l|}{\textbf{D. External clinician evaluation}} \\
\hline

Favorable response rate &
$\displaystyle
R_{\mathrm{clinical}}^{(r)}
=\frac{1}{N_r}\sum_{i=1}^{N_r}
\mathbb{I}[s_{ir}\in\mathcal A]
$ &
Both-reader favorable rate &
$\displaystyle
\begin{aligned}
R_{\mathrm{both}}
&=\frac{1}{N}\sum_{i=1}^{N}
\mathbb{I}[s_{i1}\in\mathcal A]\\[-1pt]
&\quad\cdot\mathbb{I}[s_{i2}\in\mathcal A]
\end{aligned}
$ \\
\hline

Exact reader agreement &
$\displaystyle
A_{\mathrm{exact}}
=\frac{1}{N}\sum_{i=1}^{N}
\mathbb{I}[s_{i1}=s_{i2}]
$ &
Weighted kappa &
$\displaystyle
\begin{gathered}
\kappa_w=1-\\[-1pt]
\frac{\sum_{a=1}^{J}\sum_{b=1}^{J}w_{ab}O_{ab}}
{\sum_{a=1}^{J}\sum_{b=1}^{J}w_{ab}E_{ab}}
\end{gathered}
$ \\
\hline

Linear weight &
$\displaystyle
w_{ab}=\frac{|a-b|}{J-1}
$ &
&
\\
\hline

\rowcolor{PanelGray}
\multicolumn{4}{|l|}{\textbf{E. Patient-centered evaluation and statistical analysis}} \\
\hline

Odds ratio &
$\displaystyle
\mathrm{OR}=\frac{ad}{bc}
$ &
95\% CI for OR &
$\displaystyle
\begin{gathered}
\mathrm{CI}_{95\%}(\mathrm{OR})=\\[-1pt]
\exp\!\left[
\ln(\mathrm{OR})\pm1.96
\sqrt{\frac1a+\frac1b+\frac1c+\frac1d}
\right]
\end{gathered}
$ \\
\hline

Cohen's $d$ &
$\displaystyle
d=\frac{\bar{x}_A-\bar{x}_B}{s_p}
$ &
Pooled variance &
$\displaystyle
s_p^2
=\frac{(n_A-1)s_A^2+(n_B-1)s_B^2}
{n_A+n_B-2}
$ \\
\hline

\end{tabular}

\vspace{3pt}
\begin{minipage}{\textwidth}
\scriptsize
\textit{Notes:}
TP, FP, and FN denote true positives, false positives,
and false negatives, respectively.
$K_s=10$ denotes the number of segmentation classes,
and $N$ the analysis-specific sample size.
$A_k$ and $G_k$ denote predicted and reference pixel sets,
and $n_{kj}$ denotes pixel confusion-matrix counts.
For text metrics, BP denotes the brevity penalty,
$p_k$ modified $k$-gram precision,
$P_{\mathrm L}$ and $R_{\mathrm L}$ longest-common-subsequence
precision and recall, and $\mathbf v_n$ an $n$-gram
TF-IDF vector.
$M_g$ denotes matched grounding tags,
while $N_g^{\mathrm{pred}}$ and $N_g^{\mathrm{ref}}$
denote predicted and reference tag counts.
For clinician evaluation, $\mathcal A$ denotes favorable
rating categories, $J$ the number of ordinal levels,
and $O_{ab}$ and $E_{ab}$ observed and expected rating counts.
$a,b,c,d$ denote contingency-table counts.
\end{minipage}

\end{table*}

The quantitative evaluation of PeriaTrace encompassed visual localization, semantic segmentation, diagnostic report generation, evidence grounding, and clinical applicability. The mathematical definitions of the evaluation metrics and statistical measures are summarized in Table. \ref{tab:evaluation_metrics}.

\subsubsection{Detection Metrics}
Tooth localization and caries detection were evaluated through one-to-one bounding-box matching against reference annotations using IoU. Tooth localization was assessed independently of FDI identification, while joint localization and identification additionally required correct tooth identity. An IoU threshold of 0.3 was applied to caries detection, with precision, recall, and F1 score used to quantify detection performance.

\subsubsection{Semantic Segmentation Metrics}
Semantic segmentation was evaluated using class-specific Dice similarity coefficient and IoU, foreground macro averages, and overall pixel accuracy. Macro Dice and IoU were calculated across nine foreground classes, excluding background, while pixel accuracy covered all ten classes. Class-specific results were additionally reported to characterize segmentation performance across anatomical and pathological structures.

\subsubsection{Report Generation and Evidence-Grounding Metrics}
Report generation quality was evaluated against normalized, template-standardized reference reports using BLEU-1 through BLEU-4, ROUGE-L, and CIDEr-lite, with scores averaged across the evaluation cohort. Machine-readable evidence tags were excluded from linguistic evaluation and assessed separately using grounding-tag F1. Grounding performance was further stratified by tooth bounding-box, caries lesion-box, and segmentation-related tags to quantify the correspondence between diagnostic statements and their referenced visual evidence.

\subsubsection{External Clinician Evaluation Metrics}
External clinical evaluation was conducted at two medical centers to assess image-report concordance, professional report usability, patient explanation quality, and care recommendation usability. A prespecified subset additionally evaluated the helpfulness and potentially misleading correspondence of visual grounding cues. Favorable response rates were calculated according to the corresponding ordinal rating criteria, while both-reader favorable rates required paired favorable assessments for the same image. Inter-reader reliability was assessed using exact agreement and linearly weighted Cohen's kappa.

\subsubsection{Patient-Centered Evaluation and Statistical Analysis}
Patient-centered evaluation included 196 participants without medical backgrounds, assigned to two counterbalanced survey paths (109 and 87 participants). Two cases compared professional and patient-friendly reports, while two additional cases examined bounding-box and mask-based visual grounding. Objective comprehension, actionability, and localization accuracy were summarized as correct-response proportions, while five-point Likert responses were reported using means, standard deviations, and top-two-box rates. Within-case comparisons employed Fisher's exact test for binary outcomes and two-sided Mann--Whitney U tests for ordinal responses, with odds ratios and Cohen's $d$ \cite{cohen1988statistical} used to quantify effect sizes. Multiple comparisons were adjusted using the Holm procedure \cite{holm1979simple}, and 95\% confidence intervals for proportions were calculated using the Wilson score method \cite{wilson1927probable}.

\subsection{Implementation Details}
The visual prior extractor was trained on task-specific datasets for tooth localization, caries detection, and semantic segmentation. Input radiographs were converted to RGB, resized to $1024\times1024$ using bicubic interpolation, and normalized using SAM channel statistics, without random rotation or flipping. Training was conducted on two NVIDIA RTX A6000 GPUs using distributed data parallelism for 50,000 optimization steps, with a batch size of four per task per GPU. AdamW was employed with constant learning rates of $2\times10^{-5}$ for the SAM encoder and $1\times10^{-4}$ for task-specific modules, weight decay of $10^{-4}$, gradient clipping at 1.0, and automatic mixed precision. The multitask objective combines cross-entropy and Dice losses for segmentation, cross-entropy for QC, and detection losses for tooth and caries localization, as defined in Equation. \ref{eq:impl_stage1_loss}. Tooth detection uses classification, L1, and GIoU loss \cite{rezatofighi2019generalized} weights of 1, 5, and 2, respectively, with an additional QC loss weight of 0.25, while caries detection uses corresponding weights of 2, 5, and 2. Both detection heads incorporate auxiliary decoder losses weighted by 0.5, using Hungarian matching for tooth detection and greedy one-to-one matching for caries detection. The checkpoint with the best validation performance was selected for subsequent evaluation.

\begin{equation}
\begin{aligned}
\mathcal{L}_{\mathrm{S1}}
={}&\mathcal{L}_{\mathrm{seg}}^{\mathrm{CE}}
+\mathcal{L}_{\mathrm{seg}}^{\mathrm{Dice}}
+0.25\mathcal{L}_{\mathrm{QC}}^{\mathrm{CE}}\\
&+\mathcal{L}_{\mathrm{tooth}}
+\mathcal{L}_{\mathrm{caries}}.
\end{aligned}
\label{eq:impl_stage1_loss}
\end{equation}

The diagnostic vision-language model was fine-tuned from Qwen3-VL-8B using the training set, with each sample comprising the original radiograph, annotation overlay, localization priors, and reference diagnostic JSON. LoRA was applied to attention and feed-forward projections in both visual and language components, while the pretrained weights and multimodal projector remained frozen. The LoRA rank, scaling factor, and dropout were set to 16, 32, and 0.05, respectively. Training was conducted for two epochs using AdamW with an effective batch size of 32, a learning rate of $2\times10^{-5}$, weight decay of 0.01, cosine scheduling, and a warm-up ratio of 0.03. BF16 mixed precision and gradient checkpointing were enabled, with a maximum sequence length of 6,144 tokens and an image pixel limit of 229,376. The model was optimized using autoregressive token-level negative log-likelihood, as defined in Equation. \ref{eq:impl_vlm_loss}. The checkpoint achieving the best validation performance was selected for testing, with greedy decoding and a maximum of 1,024 new tokens.

\begin{equation}
\mathcal{L}_{\mathrm{SFT}}(\phi)
=-\sum_{i,t}\log p_{\phi}
\left(y_{i,t}\mid y_{i,<t},I_i,O_i,L_i\right).
\label{eq:impl_vlm_loss}
\end{equation}
where $i$ indexes training samples, $t$ denotes the token position in the reference diagnostic JSON sequence, $y_{i,t}$ is the target token, $y_{i,<t}$ represents its preceding tokens, and $\phi$ denotes the trainable LoRA parameters. The inputs $I_i$, $O_i$, and $L_i$ correspond to the original radiograph, annotation overlay, and structured localization priors, respectively.


\section*{Code availability}
All code used in this study will be released in the GitHub repository at \href{https://github.com/zhenhuanZ/PeriaTrace}{https://github.com/zhenhuanZ/PeriaTrace}. The terms governing the use, modification, and redistribution of the code are specified in the LICENSE file included in the repository. The code is intended primarily for scientific research, and commercial use may be subject to additional restrictions under the applicable license terms. Researchers using or building upon this code are requested to cite the corresponding publication and acknowledge the GitHub repository.

\section*{Data availability}
A subset of the data collected at the First Affiliated Hospital of Nankai University may be made available from the corresponding author upon reasonable request, subject to applicable ethical approval, institutional regulations, and patient privacy requirements. Access to these data is restricted to scientific research purposes. The external evaluation data from Tianjin Union Medical Center, comprising historical archived radiographs, and the Second Hospital of Tianjin Medical University are restricted to internal institutional use due to patient privacy, confidentiality, and institutional data management requirements. These external datasets are not available for public access or external distribution.

\section*{Competing Interests Statement}
The authors declare no competing interests.

\section*{Author Contributions Statement}
Z.Z. conceived the study, developed the methodology and software, performed formal analysis and validation, curated the data, prepared the visualizations, coordinated the project, and drafted, reviewed, and edited the manuscript. Y.Z. contributed to formal analysis, resource provision, investigation, data curation, and validation. J.Z. contributed to visualization, data curation, validation, and formal analysis. S.M. contributed to resource provision, data curation, and validation. Y.W. contributed to formal analysis, validation, and investigation. T.L. supervised and coordinated the project, secured funding, provided resources, contributed to investigation and validation, and reviewed and edited the manuscript.  

\section*{Acknowledgements}
This study was supported by the China National Natural Science Foundation (Grant No. 62272248) and the S\&T Program of Hebei (No. 26240301D).

\bibliographystyle{cas-model2-names}

\bibliography{cas-refs}



\end{document}